\documentclass[journal]{IEEEtran}
\usepackage[T1]{fontenc}
\usepackage{flushend}
\usepackage{amsmath}
\usepackage[ruled,linesnumbered]{algorithm2e}
\usepackage{algpseudocode} 

\usepackage{booktabs}
\usepackage{tabu}
\usepackage{bm}
\usepackage{indentfirst}
         
\usepackage{amsmath,amssymb}
\usepackage{mathrsfs}
\usepackage{cite}
\usepackage{amssymb}
\usepackage{url}
\usepackage{hyperref}
\usepackage{graphicx}
\usepackage{amsmath}
\usepackage{amssymb}
\usepackage{booktabs}
\usepackage{multirow}
\usepackage{array} 
\usepackage[table,xcdraw]{xcolor}
\begin{document}

\title{Beyond the Prior Forgery Knowledge: Mining  Critical Clues for General Face Forgery Detection}

\author{Anwei~Luo, Chenqi~Kong, Jiwu~Huang~\IEEEmembership{Fellow,~IEEE}, Yongjian~Hu~\IEEEmembership{Senior Member,~IEEE}, Xiangui~Kang~\IEEEmembership{Senior Member,~IEEE} and~Alex~C.~Kot~\IEEEmembership{Life Fellow,~IEEE}}%
        % <-this % stops a space

% \markboth{Submitted to IEEE Transactions on Information Forensics and Security}%
% {Shell \MakeLowercase{\textit{et al.}}: Bare Demo of IEEEtran.cls for IEEE Communications Society Journals}
\maketitle

\begin{abstract}
Face forgery detection is essential in combating malicious digital face attacks. Previous methods mainly rely on prior expert knowledge to capture specific forgery clues, such as noise patterns, blending boundaries, and frequency artifacts. However, these methods tend to get trapped in local optima, resulting in limited robustness and generalization capability. To address these issues, we propose a novel Critical Forgery Mining (CFM) framework, which can be flexibly assembled with various backbones to boost their generalization and robustness performance. Specifically, we first build a fine-grained triplet and suppress specific forgery traces through prior knowledge-agnostic data augmentation. Subsequently, we propose a fine-grained relation learning prototype to mine critical information in forgeries through instance and local similarity-aware losses. Moreover, we design a novel progressive learning controller to guide the model to focus on principal feature components, enabling it to learn critical forgery features in a coarse-to-fine manner. The proposed method achieves state-of-the-art forgery detection performance under various challenging evaluation settings.

\end{abstract}


\IEEEpeerreviewmaketitle
\section{Introduction}
\IEEEPARstart{T}{he} past decade has witnessed a rapid progress on face forgery techniques such as Deepfakes \cite{hm16_20}, Face2Face \cite{dfcode}, FaceSwap\cite{thies2016face2face} and NeuralTextures \cite{thies2019deferred}. With the advent of deep learning, the falsified face contents are becoming increasingly sophisticated and realistic. Even worse, non-expert attackers can handily access off-the-shelf face editing tools like Fakeapp \cite{fakeapp} and Reface \cite{reface} to generate manipulated faces with a high level of realism. This has resulted in various pressing security concerns over financial fraud, fake news, and impersonation. Powerful as the attack technique is, there is a thin line between authentic and manipulated faces that can be hardly distinguished by human naked eyes. Therefore, it is of utmost importance to develop effective detection models to counter the malicious attacks and build the integrity of digital face contents.

Early attempts at face forgery detection can be traced back to 2018 \cite{kong2022digital}. \cite{li2018ictu, yang2019exposing, li2018exposing} mainly focused on extracting handcrafted features such as the lack of eye-blinking, head-pose inconsistency, and face warping artifacts. However, these methods suffer from limited detection accuracy. Follow-up learning-based models \cite{nguyen2019capsule, cai2020drl, cai2022learning, afchar2018mesonet, tariq2020convolutional} have demonstrated outstanding detection performance in intra-dataset evaluations. But these data-driven detectors are prone to overfitting to training data, which suffer from unsatisfactory generalization performance when deployed to unforeseen domains. To address this issue, existing methods proposed to mine common forgery clues using specific prior knowledge. In \cite{gu2022exploiting, zhao2021multi, miao2023f, 10018271}, various subtle and fine-grained clues have been identified as significant evidence for face forgery detection. The experts' prior forgery knowledge, including noise patterns \cite{luo2021generalizing, masi2020two, kong2022detect}, boundary artifacts \cite{li2020face, shiohara2022detecting}, and frequency information \cite{xu2020learning, li2021frequency}, have been widely studied to improve the generalization capability. However, these specific clues can be easily targeted by an expert attacker and are not robust to various image distortions. For example, the frequency inconsistency \cite{liu2021spatial, mi2020gan} caused by up-sampling can be eliminated by adding frequency constraints \cite{durall2020watch} or conducting post-processing operations \cite{dzanic2020fourier}. On the other hand, it is challenging to empirically choose the suitable clue type for a specific face forgery detection task.
Thus, it is only natural to ask: is there a way to drive the model to mine more critical forgery clues?

\begin{figure*}[h]
\centering
\includegraphics[scale=0.42]{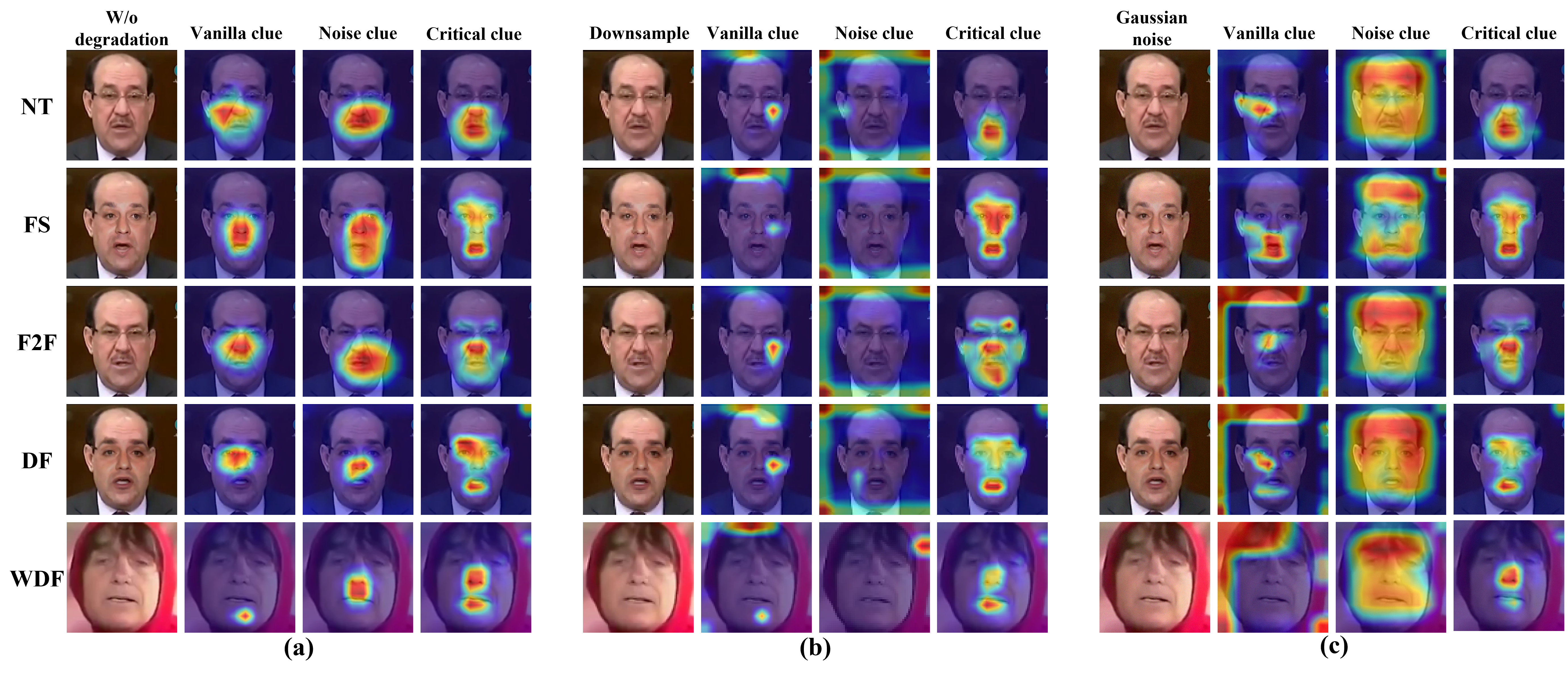}
\caption{Attention maps from different kinds of clue extractor based on the EfficientNet-B4 \cite{tan2019efficientnet} backbone. \textbf{Vanilla clue} is extracted from the vanilla CNN. \textbf{Noise clue} is extracted by suppressing the low-frequency content \cite{luo2021generalizing}. \textbf{Critical clue} is extracted by our proposed critical forgery mining (CFM) framework. The visualization in the top four rows are four types of forgery methods from Faceforensics++ \cite{rossler2019faceforensics++}, including NeuralTextures (NT) \cite{thies2019deferred}, FaceSwap (FS) \cite{FaceSwap}, Face2Face (F2F) \cite{thies2016face2face} and DeepFake (DF) \cite{DeepFake}. The last row visualizes the attention maps of the unseen dataset WildDeepfake (WDF) \cite{zi2020wilddeepfake}. We illustrate (a) non-degradation faces and two types of image degradation: (b). \emph{Downsize}, which erases high-frequency information, and (c). \emph{Gaussian Noise}, which changes the original noise pattern. The vanilla clue lacks generalization capability and is fragile to image degradations. The noise clue is sensitive to high-frequency information. In contrast, the proposed CFM framework consistently highlights the prominent manipulation regions for different types of forged images and exhibits outstanding robustness capability to image degradations.}
\label{teaser}
\end{figure*}

In this paper, our goal is to design a forgery detector which comprises the following two desired properties:
\begin{itemize}
    \item \textbf{General}: It must be general enough  in detecting manipulated faces of unseen datasets and unseen manipulation techniques. 
    \item \textbf{Robust}: It must be robust enough to detect distorted forgery faces generated by various common degradation types. 
\end{itemize}

To achieve this, we must address two key challenges: (1) preventing the model from overfitting to specific forgery clues, and (2) guiding the model to capture more critical forgery clues. In this vein, we propose to build our critical forgery mining framework from the following perspectives: (a). \textbf{Data Preparation}: We leverage prior knowledge-agnostic data augmentation to prevent the model from getting trapped in local optima and drive it to learn more generalized forgery knowledge; (b). \textbf{Learning Scheme}: We introduce a fine-grained triplet relation learning scheme that enables the model to learn more inherent feature representations; 
% (c). \textbf{Effective module}: A Manipulation Trace Enhancing (MTE) module is designed to help enhance the subtle critical forgery clues; 
(c). \textbf{Regularization Strategy}: A novel progressive learning controller (PLC) is designed to regularize the model to focus on the principal feature components but discard the less important channels; (d). \textbf{Objective Functions}:  We propose instance similarity-aware loss and local similarity-aware loss to simultaneously learn global critical features and local subtle artifacts.

Fig.~\ref{teaser} illustrates the gradient-weighted class activation mapping (Grad-CAM) \cite{selvaraju2017grad} results for the vanilla CNN, noise clue-based detector (we adopt the constraint used in \cite{luo2021generalizing}), and the proposed method across three different image types: (a). no-degradation faces; (b). downsampled faces; and (c). Gaussian noise faces. All three methods use the same EfficientNet-B4 network \cite{tan2019efficientnet} as the backbone for a fair comparison. It can be readily observed that the vanilla and noise clue-based models are vulnerable to image distortions. In turn, the proposed method can consistently focus on the critical face regions regardless of image degradation types, demonstrating its superior robustness and generalization ability for forgery detection.

The major contributions of our work can be summarised as follows:

\begin{itemize}
\item  In contrast to explore specific clues via prior forgery knowledge, we design a novel detector to mine critical
forgery clues, enabling a more general forgery face detection.

\item We propose a fine-grained triplet relation learning prototype that uses instance and local similarity-aware losses to learn general features from both global and local views. This process promotes the detector focusing on mining critical clues in both global and local manipulation regions, thereby improving the model's generalization capability.

\item We design a novel progressive learning controller (PLC), which drives the model to progressively learn forgery features in a coarse-to-fine manner. Thus, the model can effectively avoid local optima and achieve better forgery detection performance.

\item Extensive experimental results on six public datasets demonstrate that the proposed method outperforms state-of-the-art methods in terms of  robustness and generalization capability. Our visualization results further verify the effectiveness of the proposed method.

\end{itemize}

Sec. II reviews related work on prior arts in face forgery detection. Sec. III details the proposed critical forgery mining (CFM) method. Sec. IV reports comprehensive evaluation results under diverse experimental settings. Finally, Sec. V concludes this paper and discusses current limitations and possible future research directions.

\begin{figure*}[!t]
\centering
\includegraphics[width=7 in,height=3 in]{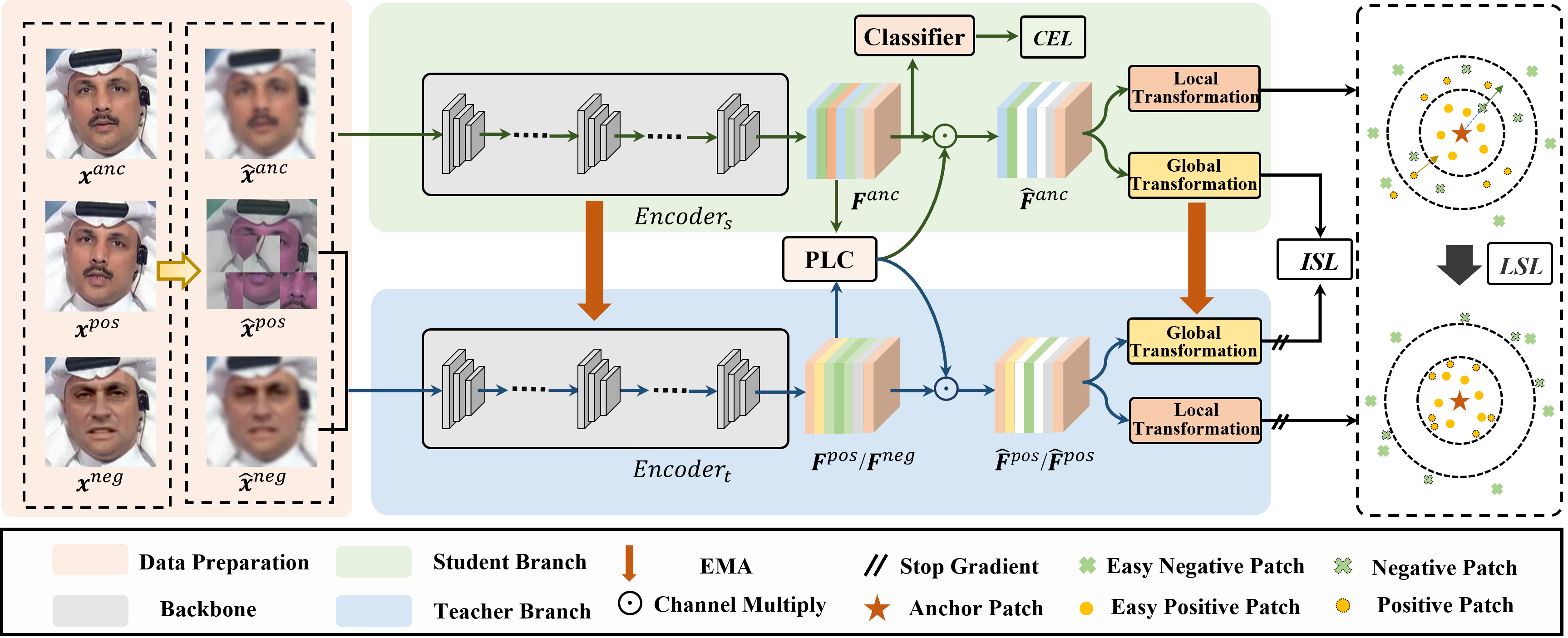}
\caption{Overview of CFM Framework. In data preparation, a fine-grained triplet is constructed while specific clues are suppressed through prior knowledge-agnostic data augmentation. The detector  learns fine-grained knowledge through Instance Similarity-aware Loss (ISL) and Local Similarity-aware Loss (LSL), which supervise the fine-grained relation learning from global and local perspectives. Furthermore, the Progressive Learning Controller (PLC) gradually controls the detector's parameter fine-tuning to capture principal information, avoiding the model's collapse to non-critical features at the early training stages.}
\label{framework}
\end{figure*}

%------------------------------Section Related work---------------------------------------
\section{Related Work}
\label{sec:relat}
% In this section, we briefly introduce the development of face forgery detection and contrastive learning. 
\subsection{Prior Knowledge-based Face Forgery Detection}
Early forged face videos tend to exhibit specific artifacts \cite{matern2019exploiting} in both spatial and temporal domains, which inspires some works to exploit hand-crafted features for forgery detection, such as lack of eye blinking \cite{li2018ictu}, inconsistency of head pose \cite{yang2019exposing}, and heart rate artifacts \cite{qi2020deeprhythm}. Powerful as deep models are, recent works take the Xception \cite{chollet2017xception} and EfficientNet \cite{tan2019efficientnet} as backbone and incorporate various types of prior knowledge to boost the detection performance \cite{zhao2021multi, li2021frequency, sun2022information, yu2022benchmarking, kong2022detect, yu2022detection, liu2022tcsd}. For instance, Face X-ray \cite{li2020face} learns to identify the boundary inconsistency left by the blending operation in forged images, while SBI \cite{shiohara2022detecting} enhances the blending artifact by synthesizing more challenging fake samples in a self-blending manner. Luo \emph{et al.} \cite{luo2021generalizing} expose the noise pattern to learn the general forgery feature. And the similar idea has been applied in MTD-Net \cite{yang2021mtd} by using the center difference convolution \cite{yu2020searching}. F$^3$Net \cite{qian2020thinking} combines global frequency information and local frequency statistics to mine frequency-aware clues. However, overemphasizing a particular type of clue can be easily countered by newly proposed forgery techniques, making them unreliable for face forgery detection. For example, these methods perform poorly when evaluated on the WildDeepfake dataset \cite{zi2020wilddeepfake} (according to the results in Table ~\ref{cross-dataset}),  which contains videos collected from the internet with high fidelity. Additionally, these methods are sensitive to various perturbations \cite{dong2022protecting} and do not meet the practical requirement in real-world applications. In turn, this paper proposes to suppress these prior knowledge-based clues and mine critical feature representations to achieve more general forgery detection.

\subsection{Face Forgery Detection via Representation Learning}
A wide variety of representation learning schemes have been proposed in general forgery detection to learn domain invariant features. Yu \emph{et al.} \cite{yu2022improving} leverage domain knowledge to learn general features across different types of forgery. FDFL \cite{li2021frequency} uses a single center loss to encourage the intra-class compactness of real face samples while relax the constraints for fake ones. The similar idea is adopted in RECCE \cite{cao2022end}, where only real images are reconstructed from their noisy versions. Lisiam \cite{wang2022lisiam} explores the robust representation by using localization invariance loss, while \cite{zhao2021learning} and \cite{chen2021local} exploit the relation between local regions to reveal the discriminative information. Additionally, RFM \cite{wang2021representative} proposes an attention-based erasing operation to encourage the model to learn features from more potential manipulation regions.

% Another interesting approach to learn representative features can be found in RFM \cite{wang2021representative}, where the authors claim that the detector easily overemphasizes local regions, and an attention-based erasing operation is proposed to encourage learning from more potential manipulation regions.

Some recent works propose using contrastive learning to improve models' generalization capability. The basic idea is to extract inherent features by maximizing mutual information between two transformation views of the same input \cite{tian2020contrastive}. This learning scheme prompts the learning of transferable representations \cite{zhang2022rethinking}. Several typical works, such as SimCLR \cite{chen2020simple}, MoCo \cite{he2020momentum}, SimSiam \cite{chen2021exploring}, and BYOL \cite{grill2020bootstrap}, have demonstrated the effectiveness of contrastive learning in various computer vision tasks. Xu \emph{et al.} \cite{xu2022supervised} have shown the feasibility of supervised contrastive learning \cite{khosla2020supervised} for face forgery detection, while Sun \emph{et al.} \cite{sun2022dual} design a dual contrastive learning framework for general representation learning. Dai \emph{et al.} \cite{dai2022attentional} propose attentional local contrastive learning to capture local forgery information. Recent works \cite{haliassos2022leveraging, zhao2022self, kong2021appearance} propose to learn consistency across different modalities, which boosts detectors to capture abundant forgery clues.
However, fine-grained features for general face forgery detection have been largely underexplored. In this paper, we carefully devise a prototype task by learning the relation in the proposed fine-grained triplet, which boosts the model to capture the subtle critical clues.

% Recently, some works propose to use contrastive learning to further improve models' generalization capability. The basic idea of contrastive learning is to capture the inherent information by maximizing the mutual information between two transformation views of the identical input \cite{tian2020contrastive}, thus prompting the learning of the transferable representation \cite{zhang2022rethinking}. Several typical works, such as SimCLR \cite{chen2020simple}, MoCo \cite{he2020momentum}, SimSiam \cite{chen2021exploring}, and BYOL \cite{grill2020bootstrap}, have demonstrated their effectiveness in various computer vision tasks. Xu \emph{et al.} \cite{xu2022supervised} verify the feasibility of the supervised contrastive learning \cite{khosla2020supervised} for face forgery detection. Sun \emph{et al.} \cite{sun2022dual} design a dual contrastive learning framework for general representation learning. Recent works \cite{haliassos2022leveraging, zhao2022self} propose to learn the consistency across different modalities, which boosts the detector to capture abundant forgery clues. Although many representative works are proposed to automatically learn meaningful features from the data, there still lacks of a deep study for the fine-grained features from the view of representation learning. To address this limitation, we carefully devise a prototype task by learning the relation in the fine-grained triplet, which actually boost model capture the subtle critical clues.  

%------------------------------Section proposed method---------------------------------------
\section{Proposed Method}
In this section, we first introduce how to construct the fine-grained triplet and provide details on how to suppress prior knowledge-based clues in Sec. III-A. Then, we present the fine-grained triplet relation learning in Sec. III-B. Finally, we introduce the designed objective functions in Sec. III-C. 

\subsection{Data Preparation}

\subsubsection{Fine-grained Triplet Construction}
% Building identity-related fine-grained triplets is convenient since the face forgery process requires information from both the source and target identities. 

As shown in Fig.~\ref{framework}, we build the fine-grained triplets using faces from forgery and the corresponding authentic videos. The anchor and positive faces are extracted from the identical real/fake video but different frames. We align each anchor and negative frames along the time dimension, ensuring the pose, expression, and background consistency between the anchor and negative data pair. The designed model brings the anchor and positive samples closer while simultaneously pushing the negative sample away from the anchor in the feature space. In this vein, the proposed fine-grained data triplet can drive the model to focus on critical forgery clues but ignore the high-level global semantic information. We denote the triplet as $(\mathbf{x}^{anc},\mathbf{x}^{pos}, \mathbf{x}^{neg})$. If $\mathbf{x}^{anc}$ is a frame from a real video, $\mathbf{x}^{pos}$ will be a different frame from the same video, and $\mathbf{x}^{neg}$ will be extracted from the corresponding manipulated video. The time interval between $\mathbf{x}^{anc}$ and $\mathbf{x}^{pos}$ is set to be more than 1 second to ensure sufficient discrepancy. We use the same sampling strategy when $\mathbf{x}^{anc}$ is from a manipulated video.

% The fine-grained triplet appears similar from a global view ($e.g.,$  backgrounds), but only shows differences in specific local regions ($e.g.,$ expressions and poses) between the anchor image and the positive/negative image. Thus, learning the relation between the fine-grained triplet can help capture the subtle clues. Let the triplet be denoted as $(\mathbf{x}^{anc},\mathbf{x}^{pos}, \mathbf{x}^{neg})$. If $\mathbf{x}^{anc}$ is a frame from a real video, $\mathbf{x}^{pos}$ is another frame from the same video, and $\mathbf{x}^{neg}$ is from the corresponding manipulated video. The time interval between $\mathbf{x}^{anc}$ and $\mathbf{x}^{pos}$ is set to be more than 1 second to ensure sufficient discrepancy. We use the same sampling strategy when $\mathbf{x}^{anc}$ is from a manipulated video.

\subsubsection{Prior Knowledge-agnostic Data Augmentation}
Many prior face forgery detection methods \cite{luo2021generalizing, li2020face, dong2022protecting} have demonstrated that specific prior knowledge ($e.g.,$ high-frequency clues, color mismatch, noise artifacts, and identity inconsistency) is effective in face forgery detection. Despite their effectiveness, these methods often rely on shortcuts, which can limit their generalization capability. In this paper, we propose a prior knowledge-agnostic data augmentation strategy to drive the model to learn more critical forgery clues and  prevent it from getting trapped in the local optima. The proposed data augmentation strategy is designed in the following aspects:  
(1) High-frequency clues: forgery videos tend to expose more forgery clues in high-frequency domain \cite{durall2020watch}. We apply \emph{Gaussian Blur}, \emph{Downscale}, and \emph{Image Compression} to mitigate the inconsistency between real and forgery faces. (2) Color mismatch: we leverage the \emph{ColorJitter} to suppress the color inconsistency between manipulated and unmanipulated regions \cite{li2020face}. (3) Noise artifacts: manipulation operations may introduce various noise patterns into forgery videos \cite{fei2022learning, luo2021generalizing}. \emph{Gaussian Noise} is applied to the data to suppress the discrepancy of different noise patterns. (4) Identity inconsistency: face swapping causes the mismatch between the inner face region and outer face region \cite{dong2022protecting, nirkin2021deepfake}. We shuffle the facial structure with \emph{Random Grid Shuffle} to remove the identity information. Note that if we use different augmentations for $\mathbf{x}^{anc}$ and $\mathbf{x}^{neg}$, the detector may learn the bias introduced by the discrepancy from augmentations rather than critical clues. Therefore, the augmentations for $\mathbf{x}^{anc}$ and $\mathbf{x}^{neg}$ are constrained to the same sampling. All prior knowledge-agnostic augmentations can be easily implemented by using the open source package Albumentations \cite{info11020125}. 

\subsection{Fine-grained Triplet Relation Learning} 
It is challenging to extract representative features from degraded inputs since the forgery clues are too subtle to mine \cite{wang2022lisiam, wang2020cnn}. In this paper, we cast the problem of detecting face forgery as a prototype learning task. This involves using fine-grained triplets as inputs to learn more critical representations.

% Previous research \cite{wang2022lisiam, wang2020cnn} has shown that accurately locating the manipulation region is challenging for detectors after the forgery clues are weakened, which suggests that the additional clues in the processed images may are difficult to mine. In our own experiments (Section IV), we also confirmed that a detector solely supervised with CE loss experienced a drop in performance. To address this issue, we designed a prototype task of fine-grained triplet relation learning, which we jointly trained with the classification task. 

After conducting data augmentation, we denote the pre-processed inputs as $\mathbf{\hat{x}}^{anc}$, $\mathbf{\hat{x}}^{pos}$, and $\mathbf{\hat{x}}^{neg}$ in Fig.~\ref{framework}. $\mathbf{\hat{x}}^{anc}$ is encoded by the student model $Encoder_{s}$, while $\mathbf{\hat{x}}^{pos}$ and $\mathbf{\hat{x}}^{neg}$ are fed forward to the teacher model $Encoder_{t}$. The parameters $\theta_{s}$ of $Encoder_{s}$ are updated with gradient back-propagation. And the parameters $\theta_{t}$ of $Encoder_{t}$ are the weighted summation of the student and teacher models using the exponential moving average (EMA), as shown in Eqn. (1) and (2).
% with an exponential hyper-parameter $\alpha$, using exponential moving average (EMA).

\begin{equation}
\left\{
             \begin{array}{lr}
             \mathbf{F} ^{\varphi} = Encoder_{s}(\mathbf{\hat{x}}^{\varphi}), & \varphi \in \left \{anc \right \}    \\
             \mathbf{F} ^{\varphi} = Encoder_{t}(\mathbf{\hat{x}}^{\varphi}), & \varphi  \in \left \{pos,neg \right \}  
             \end{array}
\right.
\end{equation}
\begin{equation}
    \theta_{t} = \alpha \theta_{t} + (1-\alpha)\theta_{s}.
\end{equation}
where $\mathbf{F} ^{\varphi}\in \mathbb{R} ^{C\times H\times W}$ is the encoded feature. By using temporal ensembling and updating the teacher parameters, the teacher model is able to produce better intermediate representations, resulting in more reliable representation learning for the student model\cite{tarvainen2017mean}.

% After encoding, we get encoded triplet denoted as $\mathbf{F} ^{\varphi}\in \mathbb{R} ^{C\times h\times w}$, among which $C$ is the channel number. By using global average pooling $GAP$ on $\mathbf{F} ^{\varphi }$, we get its corresponding channel-wise feature response  ${V} ^{\varphi }=GAP(\mathbf{F} ^{\varphi })\in\mathbb{R}^{C}$.
% $\left \{encoder_{s}(\mathbf{X}^{anc}), encoder_{t} (\mathbf{X}^{pos}),  encoder_{t} (\mathbf{X}^{neg})\right \}$
\begin{figure}[t]
  \centering
   \includegraphics[width=3.5 in]{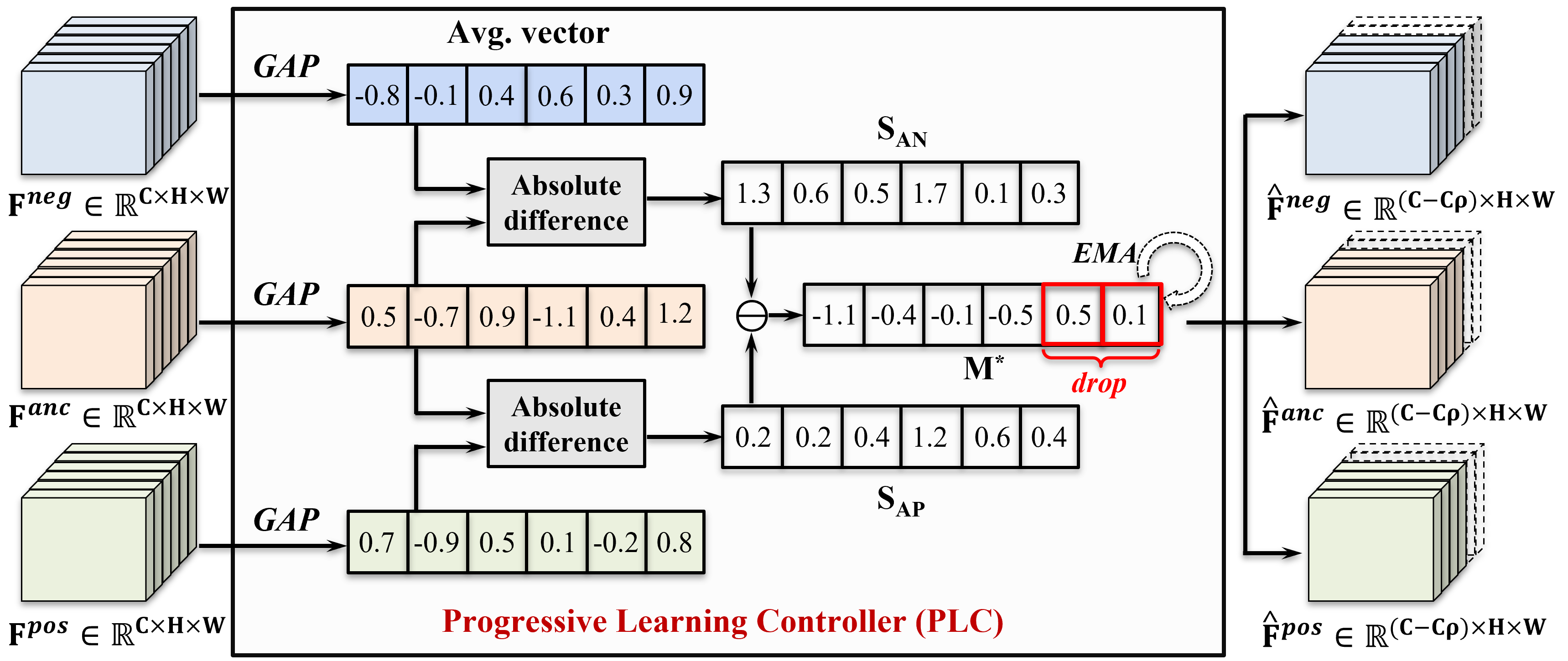}

   \caption{Pipeline of PLC. GAP denotes global average pooling and $\rho$ represents the drop rate. To determine the importance of each channel for fine-grained triplet relation learning, we calculate the channel's significance using Eqs. (3)-(6). This information is then leveraged to guide the dropout of less critical feature channels.}
   \label{PLC}
\end{figure}

\subsubsection{Progressive Learning Controller} 
Generally speaking, face forgery detection models  always performs poorly at the early training stage since it may capture biased information instead of the crucial clues. With the coarse information, the training process may collapse into local optima, leading to limited generalization capability \cite{shi2020informative}. Therefore, we design a novel progressive learning controller (PLC) to drive the detector to learn the most prominent information but discard the less important feature channels at the beginning learning stage. This ensures that the gradient contributions come from principal features. The dropped channels will be gradually added back to the features to explore more fine-grained information.

Fig.~\ref{PLC} illustrates the pipeline of PLC. We first apply global average pooling (GAP) to $\mathbf{F}^{\varphi}$. Each element in the resulted global vector can be regarded as the activation response of one type of forensics features. We hope that the model can learn critical forgery information from more discriminative features. Thus, we encourage activation responses of $\mathbf{F}^{anc}$ and $\mathbf{F}^{pos}$ to be closer, meanwhile push away the responses of $\mathbf{F}^{anc}$ and $\mathbf{F}^{neg}$. As such, we propose a new metric to measure each channel's importance, as presented in Eqn. (3):

% After applying global average pooling (GAP) to $\mathbf{F} ^{\varphi}$, each point in the resulting global vector represents the of response strength of the corresponding high-level feature. Normally, the response strength of corresponding high-level features should be similar for images of the same category, while images of different categories should have dissimilar response strengths. We want the gradients from the discriminative features to make major contributions to model updates at the beginning. After the model becomes powerful, it can gradually expand to more channels to explore richer features and further boost its performance. This process can be formulated as follow.
\begin{equation}
    \mathbf{M} = \mathbf{S}_{AP}-\mathbf{S}_{AN},
\end{equation}
\begin{equation}
    \mathbf{S}_{AP} = | GAP(\mathbf{F}^{anc}) - GAP(\mathbf{F}^{pos}) |,
\end{equation}
\begin{equation}
    \mathbf{S}_{AN} = | GAP(\mathbf{F}^{anc}) - GAP(\mathbf{F}^{neg}) |,
\end{equation}

where $\mathbf{S}_{AP}$ and $\mathbf{S}_{AN}$ are feature response difference of anchor-positive pair and anchor-negative pair, each point in $\mathbf{M}$ reflects the importance of the corresponding feature. Similarly, we use EMA updating to ensure a stable iteration for vector $\mathbf{M}^{*}$.
\begin{equation}
    \mathbf{M}^{*} = \beta \mathbf{M}^{*} + (1-\beta)\overline{\mathbf{M}},
\end{equation}
where $\beta$ is a hyper-parameter, $\overline{\mathbf{M}}$ is the average result in the current mini-batch. After sorting $\mathbf{M}^{*}$ in ascending order, the last features will be considered as unimportant information. We drop out the last $\left \lfloor C*\rho  \right \rfloor $ feature maps in $\mathbf{F} ^{\varphi }$, where $\rho$ is the drop ratio defined as:
\begin{equation}
    \rho =0.5\cos (\pi \times  \frac{E_{cur} }{E_{total} } )\times [ E_{cur}<\frac{E_{total}}{2}],
\end{equation}
where $E_{cur}$ and $E_{total}$ represent current epoch number and total epoch number, respectively. $[\cdot] $ indicates the Iverson bracket. Finally, the modified feature maps $\hat{\mathbf{F}}^{\varphi }$ will be scaled by $1/(1-\rho) $ to make the feature
unbiased.      

The designed PLC comprises the following benefits: (1). the progressive training strategy can guide the model to learn critical forgery features in a coarse-to-fine fashion, making the training process more stable; (2). it can effectively prevent the model from falling into local optima, thereby boosting its generalization.

\subsection{Objective Functions} 
Herein, we combine instance similarity-aware loss (ISL) and local similarity-aware loss (LSL) to jointly supervise the training process of the framework, further improving the general face forgery detection performance.
% promote mining critical information in fine-grained triplets
%  to learn the subtle informative feature
\subsubsection{Instance Similarity-aware Loss} $L_{ins}$ is an instance-level metric objective designed to promote mining critical information in fine-grained triplets through constraining intra-class compactness and inter-class separability. This process is formulated as follow.
\begin{equation}
    L_{ins}=max(d _{ins}+\mathbf{Z}^{anc}_{g}\cdot \mathbf{Z}^{neg}_{g}-\mathbf{Z}^{anc}_{g}\cdot \mathbf{Z}^{pos}_{g},0),
\end{equation}
where $d _{ins}$ represents the margin distance. $\mathbf{Z}^{anc}_{g}$, $\mathbf{Z}^{pos}_{g}$ and $\mathbf{Z}^{neg}_{g}$ are the mapped feature vector with a low compact dimension $C^{*}$, and they are calculated as: 
\begin{equation}
\left\{
             \begin{array}{lr}
             \mathbf{Z}^{\varphi }_{g} =norm(\hbar_{t}(GAP(\hat{\mathbf{F}}^{\varphi }))), & \varphi \in \left \{pos,neg \right \}    \\
             \mathbf{Z}^{\varphi }_{g} =norm(\hbar_{s}(GAP(\hat{\mathbf{F}}^{\varphi }))), & \varphi  \in \left \{anc \right \}  
             \end{array}
\right.
\end{equation}
where $norm( \cdot)$ refers to $L_{2}$ normalization. $\hbar_{s}$ refers to the mapping module with two linear layers, while $\hbar_{t}$ is the EMA vesion of $\hbar_{s}$. The updating of the parameters in $\hbar_{t}$ is the same as Eqn. (2).

As shown in Eqn. (9), $L_{ins}$ encourages the similarity between the anchor and positive pair is greater than that between the anchor and negative pair by a margin of $d _{ins}$. With this supervision, the detector is constrained to learn invariant critical features between anchor-positive pairs with different transformations.
Additionally, the model is able to leverage these essential differences to conduct forgery detection, even in cases where anchor-negative pair has undergone the same augmentation.

\begin{figure}[t]
  \centering
   \includegraphics[width=3.2 in]{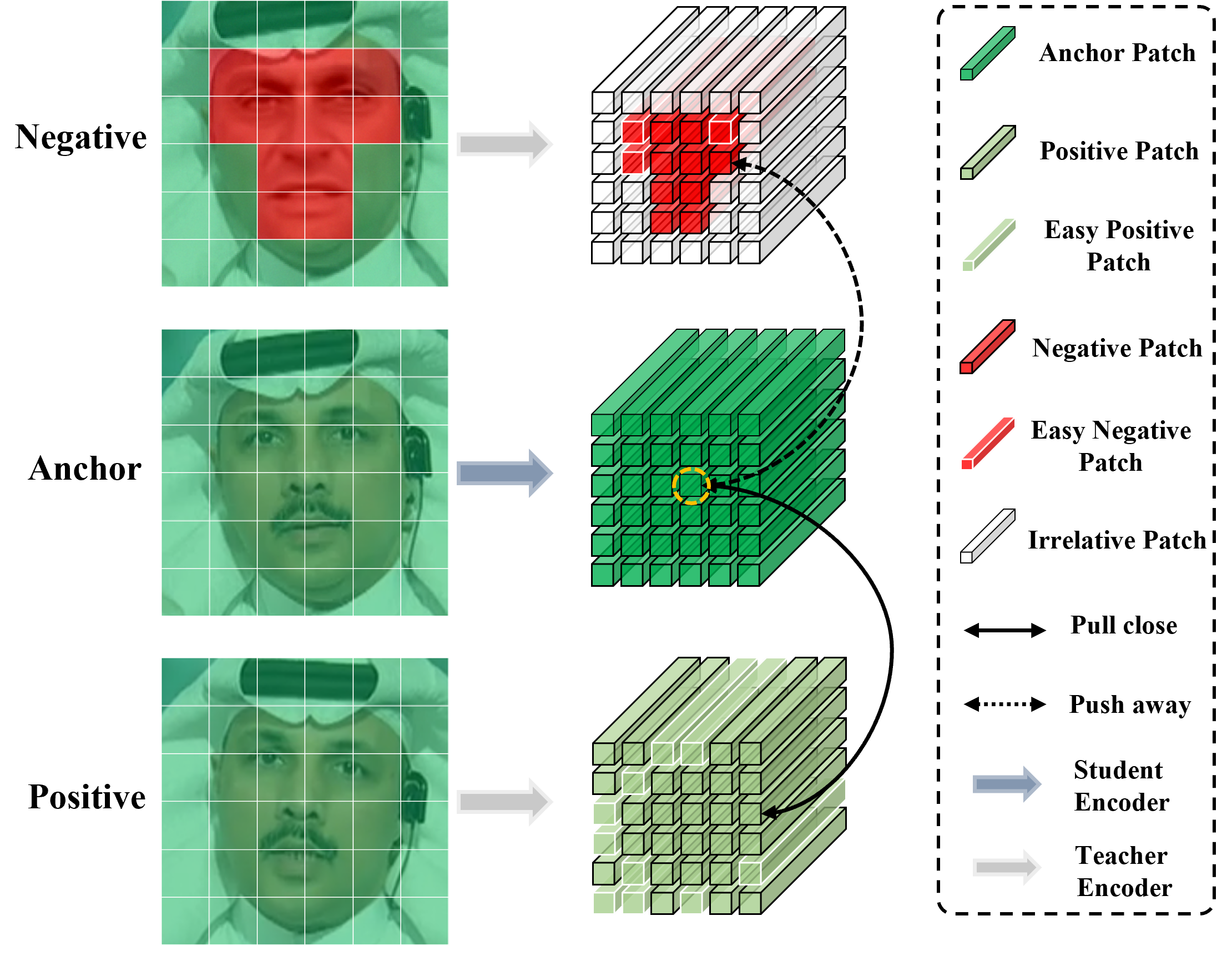}

   \caption{Illustration of $L_{loc}$. The manipulation patches (marked in red) in the negative image refers to the negative patches. Each anchor patch keeps the similarity to the positive patch larger than $s_{pos}$, whereas the similarity to the negative patch is less than $s_{neg}$. Note that we omit the augmentation for better visualization.}
   \label{LCML}
\end{figure}

\subsubsection{Local Similarity-aware Loss} 
In contrast to instance-level loss $L_{ins}$ that aims to model triplet relations in semantic-level, the local similarity-aware loss $L_{loc}$ complementarily explores the local relationship of different patches. This allows the model to learn subtle informative features. To this end, we employ the absolute pixel-level difference mask to assign local real/fake labels to each patch in the anchor, positive, and negative images, following \cite{sun2022dual,chen2021local}.
% and then we resize it into the same spatial size as $\mathbf{F} ^{\varphi}$ (i.e., $H\times W$). 
We employ $1\times1$ convolution layers to reduce the number of channels in $\mathbf{F} ^{\varphi}$ and obtain a low-dimensional embedding space $\mathbf{Z}^{\varphi }_{l} \in \mathbb{R} ^{C^{*}\times H\times W}$. Using the pixel-level difference mask, we can segment $\mathbf{Z}^{\varphi }_{l}$ into real patches $ {r}^{\varphi}_{m}$ with $m\in [1,\dots ,M]$, and fake patches ${f}^{\varphi}_{n}$ with $n\in [1,\dots ,N]$, where $M+N=HW$. More details regarding $L_{loc}$ are illustrated in Fig.~\ref{LCML}.

For a given anchor patch, patches with the same real/fake labels in the positive image are defined as positive patches, while patches with different labels in the negative image are regarded as negative patches. During the training process, we encourage the similarity between anchor and positive patches while penalize the similarity between anchor and negative patch pairs. This rule is applied to each anchor patch to enhance the model's detection performance. 
Moreover, inspired by \cite{suh2019stochastic, lin2017focal}, we further filter out the easy examples and emphasize hard examples to facilitate the model to learn more critical information. For the $i$th real label patch $r^{anc}_{i}$ , patch-level loss $ L^{i}_{loc}$ is calculated as: 
% Thus, to bring more gradient contribution for LCML and mine more general features, we ignore the contribution from the easy examples and give more weight to hard examples. 

\begin{equation}
 L^{i}_{loc} = total\_{}sim^{pos}_{i}+total\_{}sim^{neg}_{i},
\end{equation}
\begin{equation}
  total\_{}sim^{pos}_{i} = \frac{\sum_{m}(1-sim^{pos}_{i,m}) * [sim^{pos}_{i,m}<s_{pos}] * {\tau}_{i,m}}{\sum_{m} [sim^{pos}_{i,m}<s_{pos}]*{\tau}_{i,m}},
\end{equation}
\begin{equation}
  total\_{}sim^{neg}_{i} = \frac{\sum_{n}-(1-sim^{neg}_{i,n})* [sim^{neg}_{i,n}>s_{neg}] * {\tau}_{i,n}}{\sum_{n} [sim^{neg}_{i,n}>s_{neg}] * {\tau}_{i,n}}, 
\end{equation}
\begin{equation}
  {\tau}_{i,m} = 10^{(s_{pos}-sim^{pos}_{i,m})}, 
\end{equation}
\begin{equation}
  {\tau}_{i,n} = 10^{(sim^{neg}_{i,n}-s_{neg})}, 
\end{equation}
where $sim^{pos}_{i,m}= r^{anc}_{i}\cdot r^{pos}_{m}$ represents the similarity between $r^{anc}_{i}$ and its $m$th positive patch $r^{pos}_{m}$, while $sim^{neg}_{i,n}= r^{anc}_{i}\cdot f^{neg}_{n}$ represents the similarity between $r^{anc}_{i}$ and its $n$th negative patch $f^{neg}_{n}$. ${\tau}_{i,m}$ and ${\tau}_{i,n}$ are coefficients that emphasize hard examples. $s_{pos}$ and $s_{neg}$ are pre-defined thresholds to filter out the easy examples. For a fake-label patch in anchor image, we use the similar way to calculate its $ L^{i}_{loc}$,
Finally, we average the result from each anchor patch to obtain $L_{loc}$.
\begin{equation}
  L_{loc}= \frac{ {\sum_{i} L^{i}_{loc}}}{ {\sum_{i} 1}}. 
\end{equation}
% Overall, $L_{loc}$ ensures that the similarity of the positive pair is greater than $s_{pos}$ and the similarity of the negative pair is less than $s_{neg}$. The proposed local similarity-aware loss guarantees a certain margin between real and fake patches in the embedding space, thereby contributing to the forgery detection performance.

Our proposed local similarity-aware loss aims to ensure that the similarity of the positive patch pair is greater than a predefined threshold $s_{pos}$, while the similarity of the negative patch pair is less than a predefined threshold $s_{neg}$. By enforcing a margin $|s_{pos}-s_{neg}|$ between real and fake patches in the embedding space, $L_{loc}$ improves the model's forgery detection performance. Specifically, $L_{loc}$ enables the model to learn critical features that can distinguish real and fake patches, even in the presence of challenging transformations and manipulations.

\subsubsection{Total Loss}
The framework is trained in an end-to-end manner and supervised by the cross-entropy loss between the prediction result {$\hat{y}$} and ground truth label {$y$}:
\begin{equation}
    L_{ce}= -y\log_{}{\hat{y}}-(1-y)\log_{}{(1-\hat{y})},
\end{equation} 
where the label $y$ is 0 for real faces, otherwise $y$ is 1. The overall objective function consists of three components:
\begin{equation}
  L_{total}= L_{ce} + L_{ins} + L_{loc}. 
\end{equation}

In the test phase, we take the output from the classifier as the final prediction.

%%%%%%%%%%%%%%%%%%%%%%%%%%%%%%%%%%%%%%%%%%%%%%%%%%%%%%%%%%%%%%%%%%%%%%%%%%%%%%%%%%%%%%%%%%%%%%%%%%%%%%%%%%%%%%%%%%%%%%%%%%%%%%%%%%%%%%%%%%%%%%%%%%%%%%%%%%%%%%%%%%%%%%%%%%%%%%%%%
% part IV
\section{Experiments}

\begin{table}[]
 \centering
  \caption{Intra-dataset evaluation on FF++. '*' indicates the trained model provided by the authors. '\dag' indicates our re-implementation using the public official code. Methods highlighted in blue denote video-level results. }
  \label{intra-dataset}
\scalebox{0.9}{\begin{tabular}{ccccccc}
\hline
\multirow{2}{*}{Method} & \multirow{2}{*}{Venue} & \multicolumn{2}{c}{FF++ (C40)}                        & \multicolumn{2}{c}{FF++ (C23)}        \\ \cline{3-6} 
 &                               & ACC                       & AUC                       & ACC                       & AUC                       \\ \hline
Face X-Ray \cite{li2020face} & CVPR 2020   & -                         & 61.60                     & -                         & 87.40                     \\
RFM \cite{wang2021representative} & CVPR 2021         & 87.06                     & 89.83                     & 95.69                     & 98.79                     \\
GFF \cite{luo2021generalizing} & CVPR 2021      & -                     & -                     & -                    & 98.36                     \\
LTW \cite{sun2021domain} & AAAI 2021   & -                         & -                     & -                         & 99.17                     \\
Lisiam \cite{wang2022lisiam} & TIFS 2022       & 87.81                     & 91.44                     & 96.51                     & 99.13                     \\
F2Trans-S \cite{miao2023f} & TIFS 2023   & 87.20                     & 89.91                     & 96.60                     & 99.24                     \\
SBI* \cite{shiohara2022detecting} & CVPR 2022         & -                         & -                         & 77.30                     & 85.89                     \\
DCL* \cite{sun2022dual} &AAAI 2022         & -                         & -                         & 96.16                     & 99.20                     \\
Xception\dag \cite{chollet2017xception} &ICCV 2019    & 86.26                     & 89.61                     & 95.49                     & 98.92                     \\
RECCE\dag \cite{cao2022end} & CVPR 2022       & \underline{90.53}                     & \underline{94.57}                     & \textbf{97.17}            & \textbf{99.33}            \\
EN-B4\dag \cite{tan2019efficientnet} & ICML 2019       & 87.42                     & 90.15                     & 96.25                     & 99.16                     \\
CFM            &Ours & \textbf{93.29}            & \textbf{96.97}            & \underline{96.93}                     & \underline{99.25}                     \\ \hline
 \color{blue}{F3Net} \cite{qian2020thinking} & ECCV 2020       & 90.43                     & 93.30                     & \underline{97.52}                     & 98.10                     \\
 \color{blue}{FDFL} \cite{li2021frequency} & CVPR2021         & \multicolumn{1}{l}{89.00} & \multicolumn{1}{l}{92.40} & \multicolumn{1}{l}{96.69} & \multicolumn{1}{l}{\underline{99.30}} \\
 \color{blue}{Two-branch} \cite{masi2020two}& ECCV 2020  & \multicolumn{1}{l}{86.34} & \multicolumn{1}{l}{86.59} & \multicolumn{1}{l}{96.30} & \multicolumn{1}{l}{98.70} \\
 \color{blue}{Lisiam} \cite{wang2022lisiam} & TIFS 2022       & 91.29                     & 94.65                     & 97.57                     & 99.52                    \\
 \color{blue}{MRL} \cite{yang2023masked} & TIFS 2023         & \multicolumn{1}{l}{\underline{91.81}} & \multicolumn{1}{l}{\underline{96.18}} & \multicolumn{1}{l}{93.82} & \multicolumn{1}{l}{98.27} \\
 \color{blue}{CFM}              &Ours       & \textbf{94.57}            & \textbf{97.87}            & \textbf{98.54}            & \textbf{99.62}            \\ \cline{1-6} 
\end{tabular}}
\end{table}

\subsection{Experimental Setups}
\subsubsection{Implementation Details}
We use MTCNN \cite{zhang2016joint} to crop face regions and resize them to $320 \times 320$. We only sample 20 frames per video for training. Our method is implemented with PyTorch \cite{paszke2019pytorch} library on a single NVIDIA GTX3090. EfficientNet-B4 (EN-B4) \cite{tan2019efficientnet} pretrained on the ImageNet \cite{deng2009imagenet} is used as our backbone. The model is optimized by Adam \cite{kingma2014adam} with an initial learning rate of $1\times10^{-3}$, a weight decay of $1\times10^{-5}$, and a batch size of 32. The total training epoch number is 30, and the learning rate decays by 0.5 every 5 epochs. $\alpha$ and $\beta$ are set as 0.999 and 0.99, respectively. $C^{*}$ is set to 128. The selection of $d_{ins}$, $s_{pos}$, and $s_{neg}$ will be discussed in the following ablation experiments. 

\subsubsection{Evaluation Metrics}
We follow the evaluation strategy in \cite{sun2022dual}, the Accuracy (ACC), Area Under the receiver operating characteristic Curve (AUC) and Equal Error Rate (EER) are used as our evaluation metrics. We report our image-level results by default. In the video-level detection, we average the image-level scores across all frames as the final video-level score.

% \usepackage{multirow}
\begin{table*}[]
  \caption{Cross-dataset evaluation on five unseen datasets. '*' indicates the trained model provided by the authors. '\dag' indicates our re-implementation using the public official code. Methods highlighted in blue denote video-level results.}
  \label{cross-dataset}
\centering
% \resizebox{\linewidth}{!}{
\begin{tabular}{cccccccccccccc}
\hline
\multirow{2}{*}{Method} & \multirow{2}{*}{Venue} & \multicolumn{2}{c}{CDF}         & \multicolumn{2}{c}{WDF}         & \multicolumn{2}{c}{DFDC}        & \multicolumn{2}{c}{DFD}         & \multicolumn{2}{c}{DFR}  & \multicolumn{2}{c}{Avgerage}   \\ \cline{3-14} 
                      &                         & AUC            & EER            & AUC            & EER            & AUC            & EER            & AUC            & EER            & AUC            & EER             & AUC            & EER\\ \hline
Face X-ray \cite{li2020face} & CVPR 2020  & 74.20          & -              & -              & -              & 70.00          &                & 85.60          & -              & -              & -     & -    & -     \\
GFF \cite{luo2021generalizing} & CVPR 2021         & 75.31          & 32.48          & 66.51          & 41.52          & 71.58          & 34.77          & 85.51          & 25.64          & -              & -    & -      & -    \\
LTW \cite{sun2021domain} & AAAI 2021         & 77.14          & 29.34          & 67.12          & 39.22          & 74.58          & 33.81          & 88.56          & 20.57          & -              & -     & -     & -    \\
F2Trans-S \cite{miao2023f} & TIFS 2023   & 80.72          & -              & -              & -              & 71.71          & -              & -              & -              & -              & -    & -       & -   \\
SBI* \cite{shiohara2022detecting} & CVPR 2022         & \underline{81.33}          & 26.94          & 67.22          & 38.85          & \textbf{79.87} & \textbf{28.26} & 77.37          & 30.18          & 81.92          & 26.24    & 77.42     & 30.09 \\
DCL* \cite{sun2022dual} & AAAI 2022         & 81.05          & \underline{26.76}          & 72.95          & 35.73          & 71.49          & 35.90          & \underline{89.20}          & 19.46          & 89.23          & 17.96    & \underline{80.87}     & \underline{27.09} \\
Xception\dag \cite{chollet2017xception} & ICCV 2019    & 64.14         & 39.77          & 68.90          & 38.67          & 69.56          & 36.94      & 84.31       & 25.00    & 85.73   & 22.29    & 74.52          & 32.53   \\
RECCE\dag \cite{cao2022end} & CVPR 2022       & 61.42          & 41.71          & \underline{74.38}          & \underline{32.64}          & 64.08          & 40.04          & 83.35          & 24.57          & \underline{90.18}          & \underline{17.18}    & 74.68     & 31.22      \\
EN-B4\dag \cite{tan2019efficientnet} & ICML 2019       & 65.24          & 39.41          & 67.89          & 37.21          & 67.96          & 37.60          & 88.67          & \underline{18.46}          & 89.65          & 17.92    & 75.88     & 30.12      \\
CFM          & Ours          & \textbf{82.78} & \textbf{24.74} & \textbf{78.39} & \textbf{30.79} & \underline{75.82}          & \underline{31.67}          & \textbf{91.47} & \textbf{16.80} & \textbf{91.26} & \textbf{17.00} & \textbf{83.94} & \textbf{24.20} \\ \hline
\color{blue}{Lisiam} \cite{wang2022lisiam} & TIFS 2022      & 78.21          & -              & -              & -              & -              & -              & -              & -              & -              & -    & -      & -    \\
\color{blue}{F3Net} \cite{qian2020thinking} & ECCV 2020                  & 68.69          & -              & -              & -              & 67.45          & -              & -              & -              & -              & -     & -        & -   \\
\color{blue}{FTCN} \cite{zheng2021exploring} & ICCV 2021                     & 86.90          & -              & -              & -              & 74.00          & -              & -              & -              & -    & -     & -       & -              \\
\color{blue}{SBI}* \cite{shiohara2022detecting} & CVPR 2022                     & \underline{88.61}          & 19.41              & 70.27              & 37.63              &  \textbf{84.80}          &  \textbf{25.00}             & 82.68              & 26.72              & 86.85    & 22.00     & 82.64       & 26.15              \\
\color{blue}{DCL}* \cite{sun2022dual} & AAAI 2022                     & 88.24          & \underline{19.12}              & 76.87              & 31.44              & 77.57          & 29.55              & \underline{93.91}              & \underline{14.40}              & \underline{94.42}    & 12.59     & \underline{86.20}       & \underline{21.42}              \\
\color{blue}{RECCE}\dag \cite{cao2022end} & CVPR 2022                     & 69.25          & 34.38              & \underline{76.99}              & \underline{30.49}              & 66.90          & 39.39              & 86.87              & 21.55              & 93.28    & \underline{12.45}     & 78.65       & 27.65              \\
\color{blue}{CFM}        &Ours            & \textbf{89.65} & \textbf{17.65} & \textbf{82.27} & \textbf{26.80} & \underline{80.22} & \underline{27.48} & \textbf{95.21} & \textbf{11.98} & \textbf{94.80} & \textbf{11.67}  & \textbf{88.43}  & \textbf{19.11}\\ \hline
\end{tabular}%}
\end{table*}

\subsubsection{Dataset}

\textbf{FaceForensics++} (FF++) \cite{rossler2019faceforensics++}: FF++ is a standard benchmark in face forgery detection. It includes 1,000 real videos and 4,000 fake videos, which consist of four types of face manipulation techniques: Deepfakes (DF) \cite{hm16_20}, Face2Face (F2F) \cite{dfcode}, FaceSwap (FS) \cite{thies2016face2face}, and NeuralTextures (NT) \cite{thies2019deferred}. Besides, each video in FF++ has three quality levels: raw, high-quality (C23), and low-quality (C40) data. In our experiments, we consider the C23 and C40 versions to accommodate practical applications.

\textbf{Celeb-DF-v2} (CDF) \cite{li2020celeb}: CDF is a large-scale public dataset that includes 590 real videos and 5,639 fake videos. The fake videos in CDF have better visual quality than previous datasets, making it more challenging for detection.

\textbf{WildDeepfake} (WDF) \cite{zi2020wilddeepfake}: WDF is a real-world dataset with 3,805 real face sequences and 3,509 fake face sequences. All data are collected from the internet, and they include diverse scenes and forgery methods. The evaluation results on WDF reflect the detector's performance in real world scenarios.

\textbf{Deepfake Detection Challenge} (DFDC) \cite{ dolhansky2019deepfake}: DFDC is a challenging face swap dataset to evaluate detectors' performance in different scenarios. It includes 1,133 real videos and 4,080 fake videos generated by Deepfake, GAN-based, and traditional face-swapping methods.

\textbf{DeepFakeDetection} (DFD) \cite{dfd.org}: DFD is a Deepfake dataset produced by Google. It contains over 3000 manipulated videos from 28 actors in various scenes with high-level of visual realism.

\textbf{DeeperForenics-1.0} (DFR) \cite{jiang2020deeperforensics}: DFR is created by using real videos from FF++ with a newly proposed end-to-end face swapping framework. Additionally, extensive common perturbations were also applied to simulate real-world application scenarios.

For DFD, we use all the videos to evaluate the generalization performance. For the other five datasets, we follow the official meta files to 
split the corresponding datasets.

%%%%%%%%%%%%%%%%%%%%%%%%%%%%%cross-manipulation evaluation results%%%%%%%%%%%%%%%%%%%%%%%%%%%%%%%%%%%%%%%%%%%%%%%%%%%%%
\begin{table}[]
\caption{Image-level cross-manipulation evaluation results (AUC). Cross Avg. represents the average results on three cross-manipulation evaluation trials. Grey background indicates intra-manipulation results.}
\label{cross-manipulation}
% \resizebox{\linewidth}{!}{
\begin{tabular}{c|c|ccccc} 
\hline
Methods & Train                 & DF                                     & F2F                                    & FS                                     & NT                                     & Cross Avg.  \\ \hline
EN-B4 \cite{tan2019efficientnet}   &                       & \cellcolor[HTML]{E0DBDB}99.90          & 68.27                                  & 45.24                                  & 65.66                                  & 59.72          \\
RECCE \cite{cao2022end}   &                       & \cellcolor[HTML]{E0DBDB}\textbf{99.95} & 69.75                                  & 54.72                                  & \textbf{77.15}                         & 67.21          \\
CFM    & \multirow{-3}{*}{DF}  & \cellcolor[HTML]{E0DBDB}99.93          & \textbf{77.56}                         & \textbf{54.94}                         & 75.04                                  & \textbf{69.18} \\ \hline
EN-B4 \cite{tan2019efficientnet}   &                       & 81.78                                  & \cellcolor[HTML]{E0DBDB}99.13          & 58.26                                  & 66.49                                  & 68.84          \\
RECCE \cite{cao2022end}   &                       & 71.55                                  & \cellcolor[HTML]{E0DBDB}99.20          & 50.02                                  & \textbf{72.27}                         & 64.61          \\
CFM    & \multirow{-3}{*}{F2F} & \textbf{81.85}                         & \cellcolor[HTML]{E0DBDB}\textbf{99.23} & \textbf{60.12}                         & 70.80                                  & \textbf{70.92} \\ \hline
EN-B4 \cite{tan2019efficientnet}   &                       & 68.23                                  & 66.92                                  & \cellcolor[HTML]{E0DBDB}99.62          & 51.21                                  & 62.12          \\
RECCE \cite{cao2022end}   &                       & 63.05                                  & 66.21                                  & \cellcolor[HTML]{E0DBDB}99.72 & \textbf{58.07}                         & 62.44          \\
CFM    & \multirow{-3}{*}{FS}  & \textbf{72.91}                         & \textbf{71.39}                         & \cellcolor[HTML]{E0DBDB}\textbf{99.85} & 51.69                                   & \textbf{65.33} \\ \hline
EN-B4 \cite{tan2019efficientnet}   &                       & 82.12                                  & 74.95                                  & 49.32                                  & \cellcolor[HTML]{E0DBDB}99.10          & 68.80          \\
RECCE \cite{cao2022end}   &                       & 72.37                                  & 64.69                                  & 51.61                                  & \cellcolor[HTML]{E0DBDB}\textbf{99.59} & 62.89          \\
CFM    & \multirow{-3}{*}{NT}  & \textbf{88.31}                         & \textbf{76.78}                         & \textbf{52.56}                         & \cellcolor[HTML]{E0DBDB}99.24          & \textbf{72.55} \\ \hline
\end{tabular}%
\end{table}
%%%%%%%%%%%%%%%%%%%%%%%%%%%%%%%%%%%%%%%%%%%%%%%%%%%%%%%%%%%%%%%%%%%%%%%%%%%%%%%%%%%%%%%%%%%%%%%%%%%%%%%%%%%%%%%%%%%%%%%%

\subsection{Intra-dataset Evaluation}
We first compare our proposed CFM with state-of-the-art methods under the intra-dataset setting. All the models are trained on FF++ (C23) or FF++ (C40) to ensure a fair comparison. The results are shown in Table~\ref{intra-dataset}, where we bold the best results and underline the second-best results. The video-level detection results are highlighted in blue. It can be observed that our CFM achieves a significant performance improvement compared with the EN-B4 baseline on FF++ (C40) dataset. Generally speaking, it is challenging to extract discriminative features from highly compressed data as compression processes tend to erase abundant forgery information. However, our proposed CFM method focuses on capturing more critical forgery clues, thus outperforming other methods by a large margin. Compared with the recent method RECCE \cite{cao2022end}, our CFM improves the AUC from 94.57\% to 96.97\% on the challenging low-quality data. 
% We also observe satisfactory improvements when comparing our results to those obtained through Lisiam (which uses pixel-level supervision), F2Trans-S (which uses transformer-like architecture), and MRL (which involves time information). 
The excellent detection performance of CFM can be attributed to the proposed fine-grained relation learning framework, which enables the detector to focus on local differences and capture fine-grained clues even on the low-quality C40 dataset. Meanwhile, it is observed that most methods can achieve satisfactory performance on FF++ (C23) dataset due to the preservation of rich forgery clues in lowly compressed videos. Our method achieves the second-best detection performance on C23 data. On the other hand, CFM is superior to prior arts in terms of video-level detection on both C23 and C40 data, demonstrating its outstanding detection performance from another point of view.

% However, as illustrated in Fig.~\ref{teaser}, it is not easy for the detector to learn the general features because some prominent forgery clues may influence the detector to opt for a shortcut solution. While these clues may seem sufficient for intra-dataset evaluation, they are not the critical ones that can ensure the detector's generalization ability. The detail will be discussed in the next subsection.

\subsection{Generalization Evaluation}
To evaluate the generalization capability, we consider two settings: cross-dataset and cross-manipulation evaluations. 
Both scenarios are prevalent in real-world practical applications.

\subsubsection{Cross-dataset Evaluation}
Table~\ref{cross-dataset} summarizes the generalization results for cross-dataset evaluation. All models are trained on the FF++ (C23) dataset for a fair comparison. Our CFM framework exhibits an average improvement of 8.06\% in AUC and 5.92\% in EER across five unknown datasets, as compared to the EN-B4 backbone. Besides, CFM achieves a superior generalization performance on all unseen datasets compared with prior arts, such as DCL \cite{sun2022dual}, RECCE\cite{cao2022end}, and F2Trans-S \cite{miao2023f}. While previous methods commonly suffer from dramatic performance drops when deployed on unseen datasets, the proposed CFM model takes advantage of critical clue learning and achieves the best average detection performance over all datasets, demonstrating its outstanding generalization capability.
% This illustrates that the previous detector learned some non-critical clues, which had negative impacts on its generalization. 
% Compared with recent methods such as DCL \cite{sun2022dual}, RECCE\cite{cao2022end}, and F2Trans-S \cite{miao2023f}, our CFM achieves a superior generalization performance on all unseen datasets. 

% performs well on these datasets, surpassing the recent work F2Trans-S by 4.11\% in terms of AUC on DFDC. When fairly compared to DCL in the same evaluation condition, our proposed CFM achieves significant improvements. 

% In the video level evaluation, despite FTCN \cite{zheng2021exploring} explores temporal coherence information to detect fake videos, our CFM method achieves a better cross-dataset detection performance. This mainly benefits from our designed critical clues mining framework and the proposed novel data augmentation strategy. 

Although FTCN \cite{zheng2021exploring} explores temporal coherence information to detect fake videos, our CFM method can achieve a better cross-dataset detection performance. This mainly benefits from our designed critical clues mining framework and the proposed novel data augmentation strategy. 

% Overall, our method boosts the detector's ability to capture inherent artifacts as clues for detecting fake forgeries, ensuring its effectiveness across different datasets. 

%%%%%%%%%%%%%%%%%%%%%%%%%%%%%%%%%% Robustness results%%%%%%%%%%%%%%%%%%%%%%%%%%%%%%%%%%%%%%%%%%%%%%
\begin{table*}[]
\caption{Robustness evaluation on seven common perturbations. (AUC score drops compared with the pristine data)}
\label{robustness evaluation}
\centering
\scalebox{1.2}{\begin{tabular}{ccccccccc}
\hline
Method    & Saturation     & Contrast       & Block            & Noise          & Blur           & Pixelate          & Compress        & Average            \\ \hline
Xception \cite{chollet2017xception}     & -1.74          & -7.51          & -3.90            & -48.16         & -13.16          & -14.61         & -25.89          & -16.42         \\
En-B4 \cite{tan2019efficientnet}     & -1.95          & -3.98          & -1.12            & -34.45         & -4.19          & -10.17         & -25.42          & -11.61         \\
RECCE \cite{cao2022end}     & -1.06          & -5.63          & -2.01            & -49.90         & -11.15         & -14.01         & -26.26          & -15.72         \\
SBI \cite{shiohara2022detecting}      & -0.75          & -2.92          & \textbf{-0.04} & -27.59         & -10.16         & -16.27         & -18.80          & -10.93         \\
DCL \cite{sun2022dual}       & \textbf{-0.24} & -3.28          & -0.17            & -9.33          & -3.54          & -8.03          & -19.86          & -6.35          \\ 
CFM & -0.97          & \textbf{-2.36} & -0.64            & \textbf{-1.01} & \textbf{-1.08} & \textbf{-1.87} & \textbf{-18.10} & \textbf{-3.71} \\ \hline
\end{tabular}}
\end{table*}

\begin{table}[]
\caption{Image-level evaluations on different backbones. Grey background indicates intra-dataset results.}
\label{dependency experiment}
\centering
\scalebox{0.9}{\begin{tabular}{cccccc}
\hline
\multirow{2}{*}{Model} & \multicolumn{2}{c}{FF++} & CDF   & WDF   \\ \cline{2-5} 
                      & ACC         & AUC        & AUC   & AUC   \\ \hline
Xception \cite{chollet2017xception}                & \cellcolor[HTML]{E0DBDB}95.49       & \cellcolor[HTML]{E0DBDB}98.92      & 64.14 & 68.90 \\
Xception + CFM                                                        & \cellcolor[HTML]{E0DBDB}\textbf{96.04}       & \cellcolor[HTML]{E0DBDB}\textbf{98.99}      & \textbf{80.14} & \textbf{75.94} \\ \hline
EN-B0 \cite{tan2019efficientnet}                         & \cellcolor[HTML]{E0DBDB}95.89       & \cellcolor[HTML]{E0DBDB}98.86      & 64.85 & 68.33 \\
EN-B0 + CFM                              & \cellcolor[HTML]{E0DBDB}\textbf{96.01}       & \cellcolor[HTML]{E0DBDB}\textbf{99.01}      & \textbf{80.41} & \textbf{78.23} \\ \hline
MobileNet-V2 \cite{sandler2018mobilenetv2}                                & \cellcolor[HTML]{E0DBDB}95.79       & \cellcolor[HTML]{E0DBDB}98.80      & 62.76 & 69.39 \\
MobileNet-V2 + CFM                    & \cellcolor[HTML]{E0DBDB}\textbf{96.22}       & \cellcolor[HTML]{E0DBDB}\textbf{98.90}      & \textbf{78.90} & \textbf{77.41} \\ \hline
\end{tabular}}
\end{table}
%%%%%%%%%%%%%%%%%%%%%%%%%%%%%%%%%%%%%%%%%%%%%%%%%%%%%%%%%%%%%%%%%%%%%%%%%%%%%%%%%%%%%%%%%%%%%%%%%%%%%%
\subsubsection{Cross-manipulation Evaluation}
We conduct cross-manipulation experiments on FF++ (C23) to evaluate the detector's generalization capability to unknown forgery techniques. We train the model on one forgery type and evaluate it on all four types (DF, F2F, FS, and NT). As shown in Table~\ref{cross-manipulation}, when the models are trained on the DF type, our proposed CFM outperforms EN-B4 in both intra- and cross-manipulation evaluations with a near 10\% Cross Avg. AUC improvement. 
% Due to the use of early Deepfake technique in FF++, numerous trivial forgery clues (such as visual artifacts) are present in its manipulated videos. By training on such data, our CFM is able to avoid the detector from learning non-critical clues, leading to significant improvements. 
Compared to the state-of-the-art method RECCE, our CFM consistently achieves better detection results under all four evaluation settings. Although RECCE achieves an excellent intra-manipulation detection result when trained on the NT manipulation type, it suffers a significant performance drop when applied to the other three manipulation types. In contrast, 
our method successfully mines more critical forgery information and thus can generalize well to unseen forgery types.
% We speculate that RECCE may tend to overfit to specific forgery clues that are not generalizable to other forgery types.

% \usepackage{multirow}
\begin{table}[]
\caption{Impacts of $d_{ins}$, $s_{pos}$ and $s_{neg}$. Image-level AUC scores are reported.}
\label{parameter}
\centering
\begin{tabular}{cccccc}
\hline
\multicolumn{3}{c}{Hyper-parameters} & \multirow{2}{*}{FF++} & \multirow{2}{*}{CDF} & \multirow{2}{*}{WDF} \\ \cline{1-3}
$s_{pos}$    & $s_{neg}$  & $d_{ins}$  &                       &                      &                      \\ \hline
1.0          &-1.0        &1.0            &99.13                  &78.44                 &76.21                      \\
0.9          &-0.75       &1.0            &99.17                  &81.38                 &77.14                      \\
0.8          &-0.5        &1.0            &\textbf{99.25}                  &\textbf{82.57}                 &\textbf{78.11}                      \\ 
0.7          &-0.25       &1.0            &99.20                  &80.61                 &77.53                      \\ \hline
0.8        &-0.5        &1.0            &99.25                  &82.57                 &78.11                      \\ 
0.8         &-0.5       &1.2            &99.25                  &\textbf{82.78}                 &\textbf{78.39}                      \\
0.8          &-0.5       &1.4            &\textbf{99.30}                  &80.65                 &76.09                      \\ \hline
\end{tabular}
\end{table}

\subsection{Robustness Evaluation}
Forged videos transmitted online always inevitably involve unknown transformations, and various forms of image degradation can erase various low-level forgery clues \cite{dong2022protecting}. To evaluate the robustness of a practical detector, we follow the protocol outlined in \cite{jiang2020deeperforensics} and incorporate seven common perturbations in this evaluation. As Gaussian noise and blur have been applied in our data preparation, we use motion blur and multiplicative noise instead. We report the AUC score drops compared with pristine data in Table~\ref{robustness evaluation}. We can observe that our CFM is robust across different perturbations.  Compared to the EN-B4 backbone, our proposed CFM achieves a 7.90\% average improvement, going from -11.61\% to -3.71\%. Overall, the CFM achieves the best robustness performance among all methods. Previous methods perform poorly on the last three perturbation types that remove high-frequency information of input data. This indicates that these methods tend to capture high-frequency forgery information, thus getting trapped in local optima. In turn, our CFM takes adavantage of the prior knowledge-agnostic data augmentation and fine-grained relation learning framework, thereby achieving more robust detection performance.

% \usepackage{multirow}
\begin{table}[]
\centering
\caption{Impacts of the Weighting strategy. Image-level AUC scores are reported.}
\label{weight}
\begin{tabular}{cccc}
\hline
Variant & FF++  & CDF   & WDF   \\ \cline{2-4} \hline
w/o weighting               & 99.18 & 80.95 & 77.08 \\
w/ weighting             & \textbf{99.25} & \textbf{82.78} & \textbf{78.39}      \\ \hline
\end{tabular}
\end{table}
\subsection{Flexibility Evaluation}
Our CFM can be flexibly applied to various backbones to improve their forgery detection performance. We evaluate the effectiveness of our CFM framework by integrating it with Xception \cite{chollet2017xception}, EfficientNet-B0 \cite{tan2019efficientnet}, and MobileNet-V2 \cite{sandler2018mobilenetv2}. We train these models on FF++ (C23) and evaluate their performance under both intra- and cross-dataset settings. The results are reported in Table~\ref{dependency experiment}. Benefiting from our proposed CFM, the performances of different models demonstrate promising improvements, especially in cross-dataset evaluations. Even for lightweight models like MobileNet-V2 and EN-B0, the generalization performances are still state-of-the-art. This demonstrates that the CFM is a universal framework to mine critical forgery clues and can be flexibly adapted to different backbones in a plug-and-play fashion.

% In this paper, EfficientNet-b4 is adopted as the standard backbone, here we further investigate the effectiveness of our CFM framework by incorporating with other backbones. We select several state-of-the-art backbones, i.e., XceptionNet \cite{chollet2017xception}, EfficientNet-b0 \cite{tan2019efficientnet} and MobileNet-V2 \cite{sandler2018mobilenetv2}, as our detectors, and train them on FF++ (HQ). We evaluate their performance on intra-domain and cross-domain datasets, which are illustrated in Table 5. We observe that our CFM framework facilitates the detector's generalization regardless of different backbones. When we use EfficientNet-b0, a small model with only 4M parameters, as the backbone in our CFM, the generalization performance on CDF dataset and WDF dataset is still better than some state-of-the-art, e.g., MAT \cite{zhao2021multi} and F3Net \cite{qian2020thinking}. In addition, we use Grad-CAM to visualize the heatmaps generated from these detectors. From Fig. 2, the decision region in different detectors can precisely locate the manipulation region. For the same forged image, their decision regions are similar. It demonstrates that our CFM is a universal framework for different backbones.  

% \usepackage{multirow}

\subsection{Ablation Experiments}
In this subsection, we conduct extensive ablation experiments to analyze the impacts of different components of the proposed CFM framework. We train all models on the FF++ (C23) set and evaluate intra-dataset performance on the FF++ (C23) test set and  cross-dataset performance on CDF and WDF datasets. Since  generalization is one of the significant issues in face forgery detection, we mainly focus on analyzing the generalization capability of different models. 

%%%%%%%%%%%%%%%%%%%%%%%%%%%%%%%%%%%%%%%%%%%%%%%%%%%%%%%%%%%%%%%%

%%%%%%%%%%%%%%%%%%%%%%%%%%%%%%%%%%%%%%%%%%%%%%%%%%%%%%%%%%%%%%%%%%%%
\subsubsection{Impacts of $d_{ins}$, $s_{pos}$ and $s_{neg}$}
For the fine-grained triplet relation learning task,  $d_{ins}$ in Eqn.~(8) controls the global distance metric in $L_{ins}$, while $s_{pos}$ and $s_{neg}$ defined in Eqn.~(11) and Eqn.~(12) are the similarity constraints for the anchor-positive and anchor-negative pairs, respectively. Firstly, we fix $d_{global}$ to 1 and analyze the parameter in ISL. Table~\ref{parameter} shows that when $s_{pos}=1$ and $s_{neg}=-1$, there are no easy examples during the training process. In this case, the more useful information from hard examples has less contribution to the model, thereby damaging the detection performance. We find the model achieves the best performance with $s_{pos}=0.8$ and $s_{neg}=-0.5$. Then, we fix the parameters in LSL and analyze $d_{ins}$ in ISL. We find that a larger value of $d_{ins}$ can achieve better intra-dataset results but a lower cross-dataset result. Finally, we set $d_{ins}$ as 1.2.

Moreover, we analyze the impacts of the weighting strategy in LSL. As illustrated in Table~\ref{weight}, we observe obvious performance improvements in both intra- and cross-dataset evaluations. This also suggests that the hard examples are crucial to lead the model to learn more discriminative features.

%%%%%%%%%%%%%%%%%%%%% Ablation each components %%%%%%%%%%%%%%%%%%%%%%%%%%%%%%
\begin{table}[]
\centering
\caption{Impacts of different components of the proposed CFM. Aug. indicates augmentation. Image-level AUC scores are reported.}
\label{componets ablation}
\begin{tabular}{cccc}
\hline
Variant           & FF++           & CDF            & WDF            \\ \hline
baseline          & 99.16          & 65.24          & 67.89          \\
baseline + Aug.   & 98.58          & 71.57          & 67.02          \\
w/o ISL          & 99.18          & 80.64          & 78.05          \\
w/o LSL          & 99.09          & 79.58          & 76.21          \\
w/o PLC           & 99.15          & 78.84          & 76.85          \\
CFM               & \textbf{99.25} & \textbf{82.78} & \textbf{78.39} \\ \hline
\end{tabular}
\end{table}

\begin{figure}[]
\centering
\includegraphics[scale=0.4]{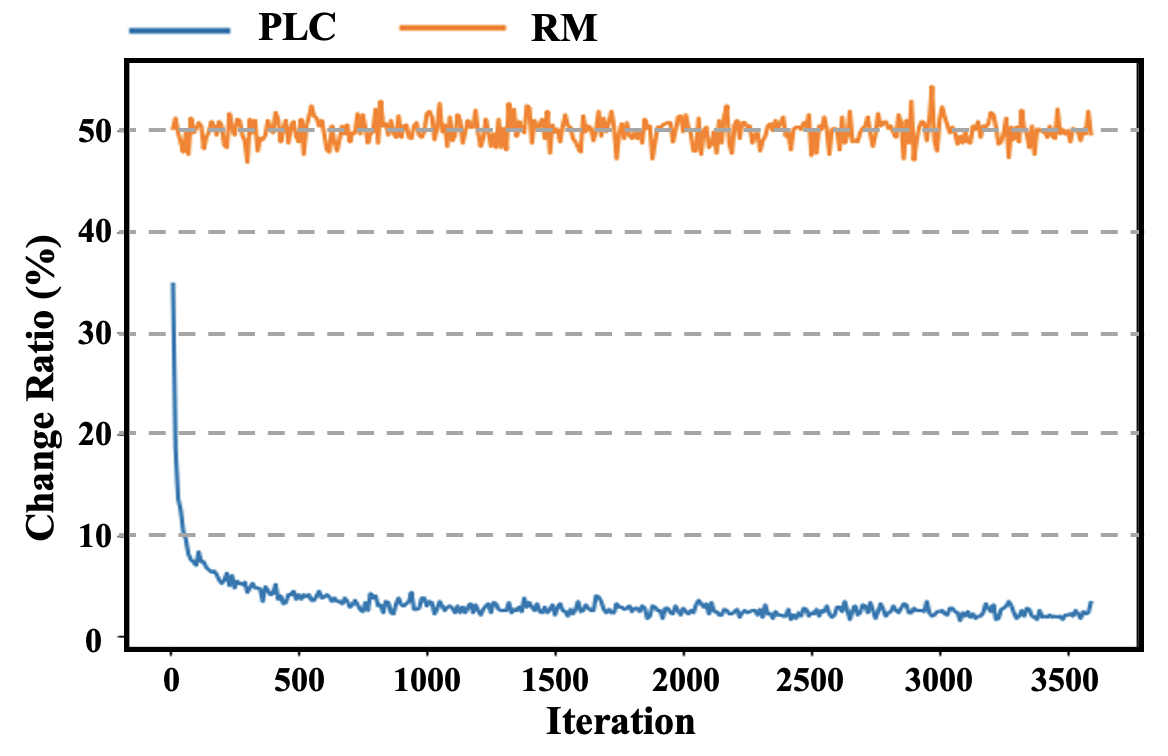}
\caption{Change ratio of the masked feature channels. 'RM' denotes random masking strategy. The change ratio refers to the ratio of the number of changes in masked feature channels between consecutive iterations to the total number of masked channels.}
\label{index}
\end{figure}

%%%%%%%%%%%%%%%%%%%%%%%%%%%%%%%%%%%%%%%%%%%%%%%%%%%%%%%%%%%%%%%%%%%%%%%%%
\subsubsection{Effectiveness of Different Proposed Components}
As shown in Table~\ref{componets ablation}, we conduct ablation experiments to verify the effectiveness of each component in our proposed CFM. Firstly, we train a baseline model using the proposed data augmentation. Although the AUC score drops a little in the intra-dataset experiment, 
the proposed prior knowledge-agnostic data augmentation boosts the generalization detection performance on the CDF dataset by a large margin (going from 65.24\% to 71.57\%).
For the ISL, LSL, and PLC components, we can observe a certain performance drop in both intra- and cross-dataset evaluations when any one of them is removed, demonstrating their effectiveness in face forgery detection. More specifically, when the PLC module is removed, the AUC drops from 82.78\% to 78.84\%. Besides, ISL and LSL drive the model to learn the discrepancy from the fine-grained triplet, leading to better detection performance. 
% All of them are the important components to learn the general features. 

%%%%%%%%%%%%%%%%%%%%%%%%%%%%%%%%%%%%%%%%%%%%%%%%%%%%%%%%%%%%%%%%%%%%%%%%%

\begin{table}[]
\centering
\caption{Generalization performance on different masking strategies. }
\label{plcablation}
\begin{tabular}{ccccc}
\hline
\multirow{2}{*}{Strategy} & \multicolumn{2}{c}{CDF}         & \multicolumn{2}{c}{WDF}         \\ \cline{2-5} 
                          & AUC            & EER            & AUC            & EER            \\ \hline
w/o regularization        & 78.84          & 27.91          & 76.85          & 31.93          \\
RM                        & 79.02          & 27.89          & 76.10          & 31.59          \\
PLC                       & \textbf{82.78} & \textbf{24.74} & \textbf{78.39} & \textbf{30.79} \\ \hline
\end{tabular}
\end{table}

\subsubsection{Impacts of Different Masking Strategies for PLC}
In this subsection, we analyze the properties of the PLC module to answer the following questions: (1). Do the masked channels vary dramatically during the training process? (2). Can similar results be achieved with random masking (RM) strategy (i.e., dropout operation)? To do this, we first train a model using the RM strategy. As shown in Fig.~\ref{index}, our results reveal that the change ratio in PLC is lower than 5\% after a few iterations, while the change ratio in RM is around 50\%. Therefore, we can conclude that the masked channels are stable in PLC. Furthermore, the results presented in Table~\ref{plcablation} demonstrate that the usage of PLC can lead to a better generalization performance. No obvious improvements are observed when using RM. This indicates that the generalization gains mainly come from the progressive learning strategy of PLC instead of the random dropout operation.

% \vspace{-0.2}
% \usepackage{multirow}
\begin{table}[]
\caption{Ablation experiments on various prior knowledge-agnostic clues. Four types of augmentations have been incorporated: H-Rel. Aug. (high-frequency clues relative augmentation), C-Rel. Aug. (color mismatch relative augmentation), N-Rel. Aug. (noise artifacts relative augmentation), and I-Rel. Aug. (identity inconsistency relative augmentation).}
\label{augmentation ablation}
\centering
\begin{tabular}{ccccc}
\hline
\multirow{2}{*}{Variant}    & \multicolumn{2}{c}{CDF}         & \multicolumn{2}{c}{WDF}         \\ \cline{2-5} 
                            & AUC            & EER            & AUC            & EER            \\ \hline
w/o Aug.            & 63.91          & 39.63          & 70.12          & 35.68          \\
w/o H-Rel. Aug. & 70.13          & 35.93          & 73.62          & 32.97          \\
w/o C-Rel. Aug. & 76.15          & 30.76          & 75.65          & 31.44          \\
w/o N-Rel. Aug. & 80.92          & 27.34          & 76.54          & 32.12          \\
w/o I-Rel. Aug. & 78.18          & 29.61          & 75.87          & 32.35          \\
CFM                         & \textbf{82.78} & \textbf{24.74} & \textbf{78.39} & \textbf{30.79} \\ \hline
\end{tabular}
\end{table}
% \vspace{-0.1cm}

% \usepackage{multirow}
\begin{table}[]
\caption{Ablation on Pairwise Constraint for Anchor-Negative Pair.}
\label{constraint}
\centering
\begin{tabular}{ccccc}
\hline
\multirow{2}{*}{Variant} & \multicolumn{2}{c}{CDF}         & \multicolumn{2}{c}{WDF}         \\ \cline{2-5} 
                         & AUC            & EER            & AUC            & EER            \\ \hline
w/o constraint           & 81.46          & 26.00          & 77.51          & 30.94          \\
w/ constraint          & \textbf{82.78} & \textbf{24.74} & \textbf{78.39} & \textbf{30.79} \\ \hline
\end{tabular}
\end{table}

\begin{figure*}[!t]
\centering
\includegraphics[scale=0.5]{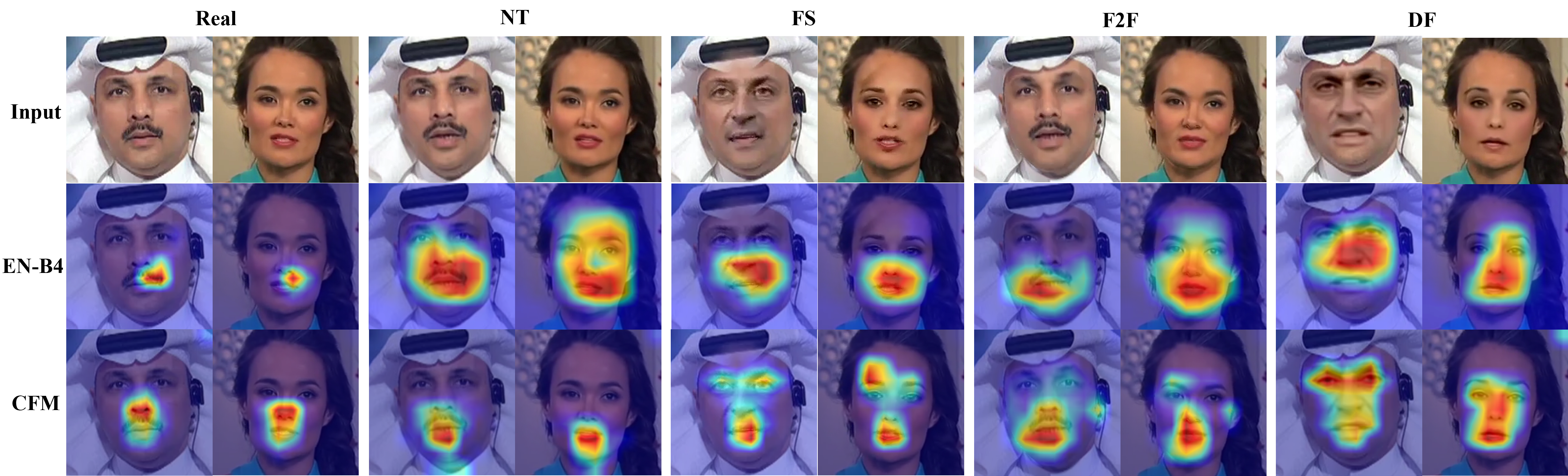}
\caption{Grad-CAM visualization of baseline model and our CFM on FF++ (intra-dataset visualization). The warmer color indicates higher confidence for model's decision.}
\label{intra_cam}
\end{figure*}

\begin{figure}[t]
  \centering
   \includegraphics[width=3.2 in]{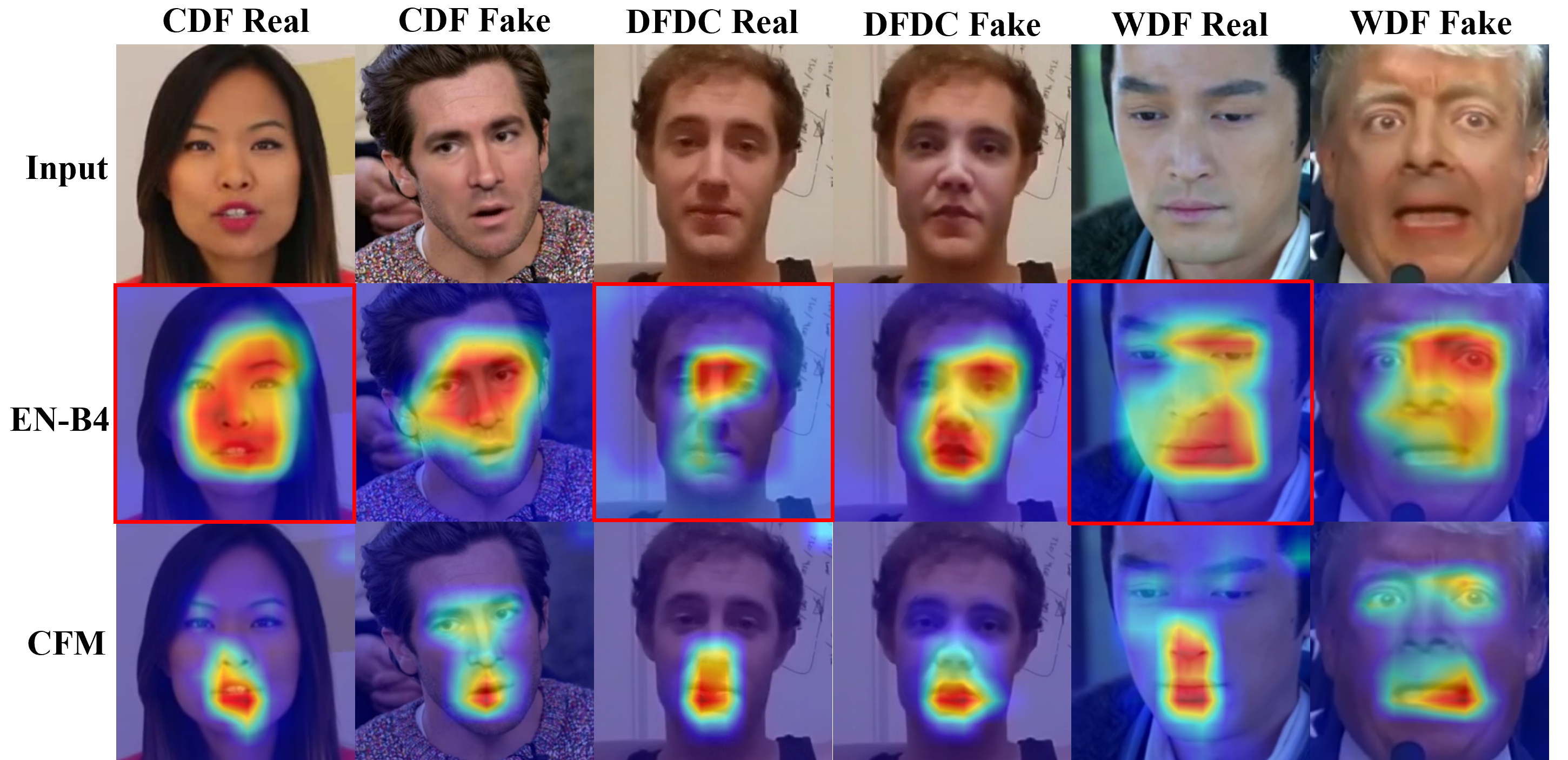}

   \caption{Grad-CAM visualization of baseline model and our CFM on CDF, DFDC and WDF (cross-dataset visualization). The red box means the detector makes a wrong prediction.}
   \label{cross_cam}
\end{figure}

\begin{figure}[t]
  \centering
   \includegraphics[width=3.2 in]{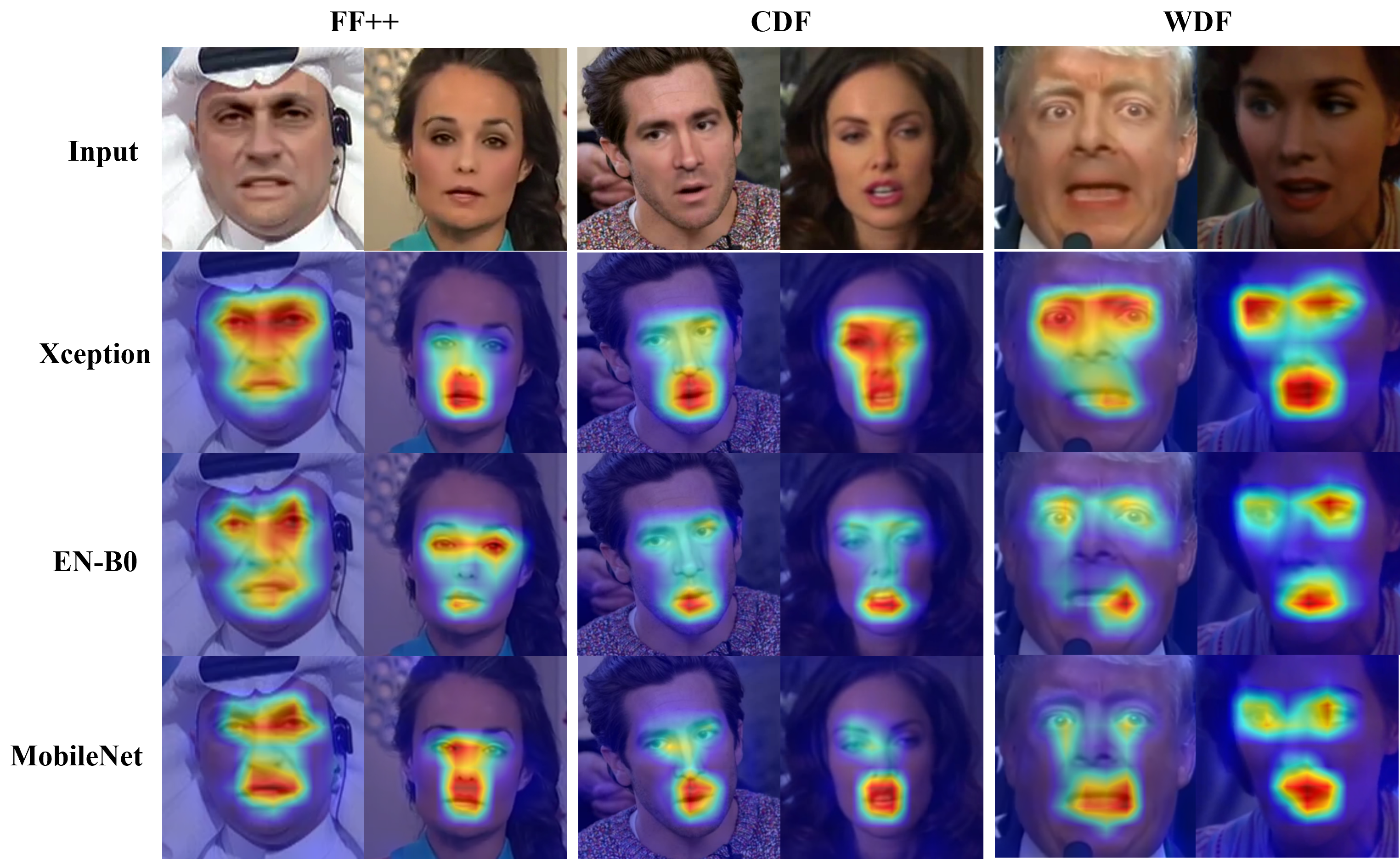}

   \caption{Grad-CAM visualization of CFM with different backbones for deepfake faces in FF++, CDF and WDF.}
   \label{multi_model_cam}
\end{figure}

\begin{figure}[t]
  \centering
   \includegraphics[width=3.2 in]{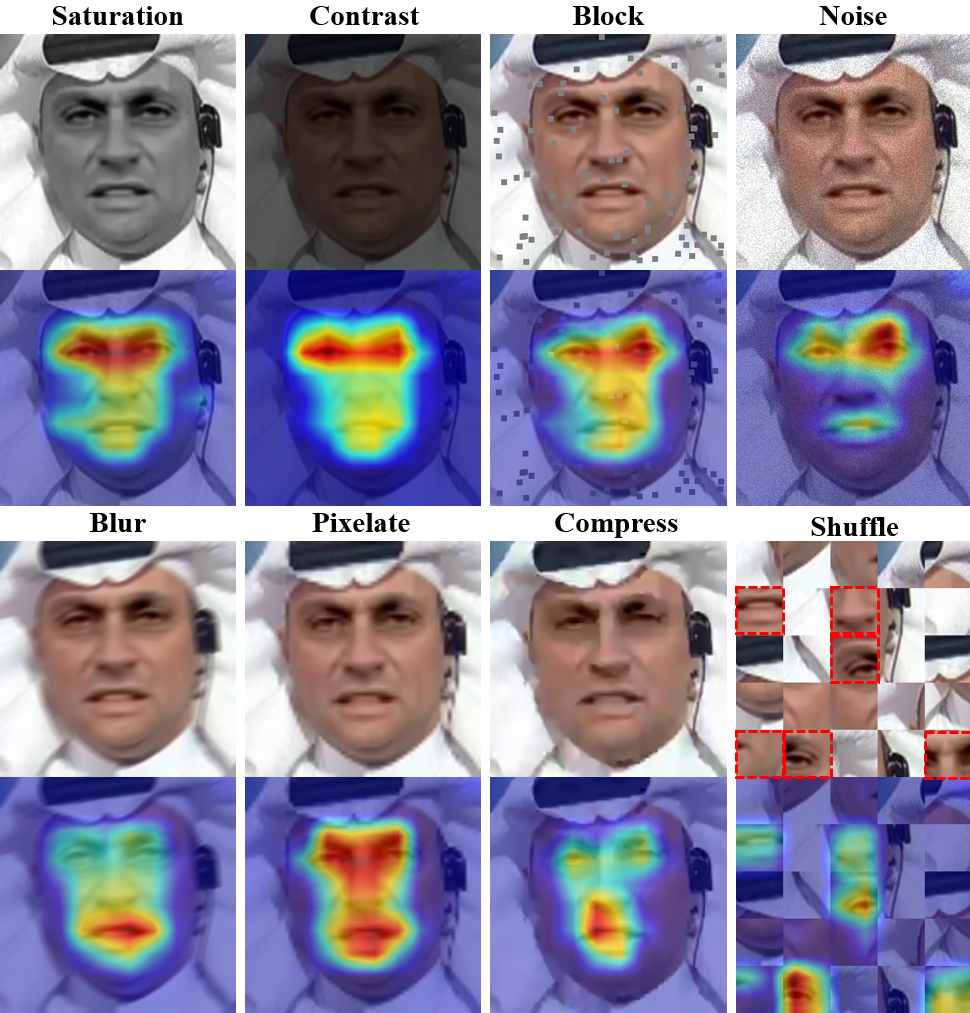}

   \caption{Grad-CAM visualization of our CFM on different common perturbations.}
   \label{robust_cam}
\end{figure}

\begin{figure}[t]
  \centering
   \includegraphics[width=3. in]{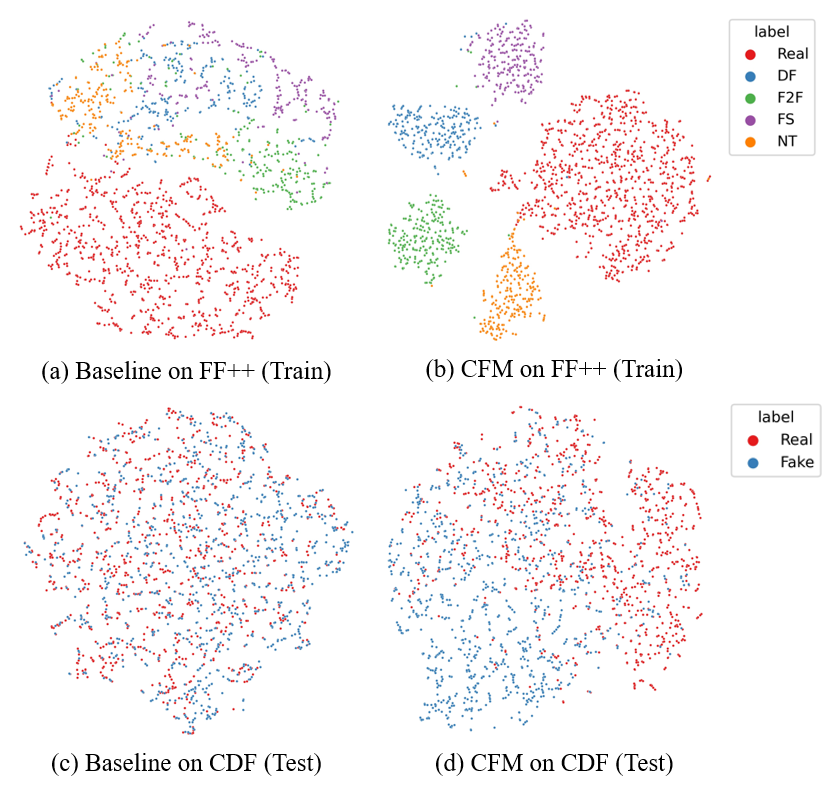}
   \caption{T-SNE feature distribution visualizations of baseline and CFM.}
   \label{tsne}
\end{figure}

\subsubsection{Impacts of the Prior Knowledge-agnostic Augmentation}
The usage of prior knowledge-agnostic augmentation is essential to prevent the model from overfitting to non-critical clues. To measure the impacts of each type of augmentation described in Sec. III-A, we train five different variants, including one model without any augmentations, and four models without one specific type of augmentation. Table~\ref{augmentation ablation} summarizes the generalization performance of each variant. From the first row, it can be readily observed that the generalization performance is poor without any augmentations. Performance degradation can be observed when any type of augmentation is removed. In the second row of Table~\ref{augmentation ablation} , without the high-frequency clues relative augmentation, the AUC scores drop by 12.65\% on CDF and 4.77\% on WDF. This ablation experiment indicates that using prior knowledge-agnostic data augmentation is crucial for mining critical forgery clues and improving the generalization performance.

\subsubsection{Effectiveness of Pairwise Constraint for Anchor-Negative data Pair}
In data preparation, we apply the same prior knowledge-agnostic augmentation to both anchor and negative images. To investigate the impact of such pairwise constraint, we conduct an ablation experiment in Table~\ref{constraint}. By using this proposed data constraint, the model is able to achieve a more generalized detection performance on challenging CDF and WDF datasets. This is because the constraint encourages the model to focus on the inherent forgery information, rather than being distracted by the augmentation itself.

% When the constraint for the anchor-negative pair is removed, the model learns  some features caused by the inconsistent augmentation to distinguish anchor and negative samples. However, with the constraint in place, the model avoids such bias and improves its generalization performance.

\subsection{Visualization}
In this subsection, we visualize attention maps of inputs and feature distributions to better illustrate the superiority of our proposed CFM.

\subsubsection{Visualization for Detector's Attention}

We use the Grad-CAM to visualize the decision regions of the detector. As shown in Fig.~\ref{intra_cam}, the heatmaps generated by CFM highlight different facial regions for real and fake faces. Specifically, the heatmaps identify the mouth region for NT and F2F and the facial feature regions for DF and FS. Since NT only modifies the mouth region and DF mainly changes the facial features, this indicates that the heatmaps produced by our CFM accurately reveal the manipulated regions. 

Fig.~\ref{cross_cam} shows the Grad-CAM visualization results in the cross-dataset evaluation. The heatmaps generated by the baseline model locate similar regions in both real and fake images. And we observe that the baseline model incorrectly identified real images as fake, indicating that the learned clues in the baseline model are not generalizable to unseen data. In contrast, the Grad-CAM maps of our CFM are more discriminative between real and fake faces, regardless of different datasets. For the real-fake data pair in the DFDC dataset, we can observe that our CFM accurately identifies the prominent differences between real and fake faces.

Fig.~\ref{multi_model_cam} shows the visualization results of our proposed CFM framework on different backbones. 
We observe that the heatmaps of CFM on different backbones mainly focus on facial feature regions, successfully localizing the face-swapping parts.
Moreover, for the same fake image, different backbones rely on similar regions to make the final decision, demonstrating the flexibility of our CFM framework from another point of view.

Fig.~\ref{robust_cam} presents the visualization results of the proposed CFM model on perturbed data. The detector successfully identifies prominent manipulation regions under various perturbations, demonstrating the robustness of our model. Even if we shuffle the patches of the input image, the highlighted regions (marked by red dotted boxes) can still reveal the manipulated areas, demonstrating that the proposed CFM can successfully mine local fine-grained forgery clues.

\subsubsection{T-SNE Feature Embedding Visualization}
We apply t-SNE \cite{van2008visualizing} to compare the baseline model with our CFM under both intra- and cross-dataset settings, as illustrated in Fig.~\ref{tsne}. The intra-dataset results demonstrate that the proposed CFM can better separate the real and fake faces in the feature space. In the cross-dataset evaluation, the CFM can achieve a more accurate classification between real and fake. Overall, the proposed CFM model exhibits better detection performance and better generalization capability, demonstrating its effectiveness in capturing more critical forgery clues.
% We plotted the t-SNE \cite{van2008visualizing} results of the baseline model and our CFM under both intra-dataset and cross-dataset conditions. As can be seen from Fig. ~\ref{tsne}, by equipping with CFM, EN-B4 learns a compact intra-class distribution and marginal inter-class distribution. When tested on the unseen dataset, our CFM shows a much more separate distribution. 

\section{Conclusion and Future work}
This paper presented a novel and effective Critical Forgery Mining (CFM) framework for face forgery detection. By using prior knowledge-agnostic data augmentation, the CFM framework successfully extracted abundant critical forgery information, improving the model's robustness and generalization capability. The fine-grained triplet learning scheme also enabled the model to learn more inherent features. Moreover, the novel progressive learning controller facilitated the detector to mine forgery features in a coarse-to-fine manner, contributing to better detection performance. Finally, we found that our instance and local similarity-aware objective functions could lead the model to learn both global critical features and subtle local artifacts, further boosting the final detection performance. Extensive quantitative and qualitative experimental results demonstrated the accuracy, robustness, and generalization capability of our CFM framework. In summary, the CFM framework provides an effective and reliable solution to the challenging face forgery detection problem. We believe the proposed method can shed light on the community and facilitate  the development of more powerful forgery detectors. 

While the proposed CFM framework is genral and robust under a wide variety of experimental settings, the temporal information has been largely under-explored in this paper. In future work, incorporating temporal features into our model and exposing temporal inconsistency in fake videos opens an important research path forward. 

\ifCLASSOPTIONcaptionsoff
  \newpage
\fi

\bibliographystyle{IEEEtran}
\bibliography{refs}

\end{document}